\documentclass[10pt,letterpaper]{article}

\usepackage[top=0.85in,left=2.75in,footskip=0.75in]{geometry}

\raggedright
\usepackage{amsmath,amssymb}
\usepackage{changepage}
\usepackage{algorithm}
\usepackage{algpseudocode}
\usepackage{float}
\usepackage{siunitx}
\usepackage{textcomp,marvosym}
\usepackage{cite}
\usepackage{nameref,hyperref}
\usepackage[right]{lineno}
\usepackage[nopatch=eqnum]{microtype}
\usepackage[table]{xcolor}
\usepackage{array}
\usepackage{graphicx}
\usepackage{epstopdf}

\newcolumntype{+}{!{\vrule width 2pt}}

\newlength\savedwidth

\usepackage[
    aboveskip=1pt,
    labelfont=bf,
    labelsep=period,
    justification=raggedright,
    singlelinecheck=off
]{caption}

\usepackage[numbers,sort&compress]{natbib}
\usepackage{lastpage,fancyhdr}

\fancyheadoffset[L]{2.25in}
\fancyfootoffset[L]{2.25in}

\begin{document}

\vspace*{0.2in}

\begin{flushleft}

{\Large
\textbf{Emergent Control of Ant Learning Walks in a Mobile Robot}
}\\

\vspace*{0.2in}

Thomas Misiek\textsuperscript{1},
Andrew Philippides\textsuperscript{1},
Paul Graham\textsuperscript{2},
James Knight\textsuperscript{1}\\

\bigskip

\textbf{1} Sussex AI, School of Engineering and Informatics,
University of Sussex, Brighton, UK\\

\textbf{2} School of Life Sciences,
University of Sussex, Brighton, UK\\

\bigskip

* \texttt{t.misiek@sussex.ac.uk}

\end{flushleft}

\section{Abstract}

Before foraging, desert ants perform learning walks (LWs) to acquire visual memories around their nest. With experience, these walks expand, become straighter, and contain fewer visual scans. We previously proposed a model in which these behaviours emerge from the angular mismatch between path integration (PI) and a mushroom-body-inspired visual compass (VC). Here, we test whether this mechanism can generate learning walks on a physical robot, despite the sensory and motor noise inherent to real-world conditions.

We implement the model on a four-wheeled robot equipped with a panoramic camera, an IMU and wheel encoders providing compass and stride information for PI, and test it outdoors using a replicated desert-ant experimental setup. The PI--VC mismatch regulates walking speed, turning, pirouettes, and returns to the nest, while PI provides a teaching signal for visual learning. The robot autonomously decides when to acquire visual memories, while its behaviour gradually shifts from exploration towards foraging as visual learning progresses.

The robot reproduces desert-ant LW features, including expanding and straightening trajectories, decreasing pirouette frequency, and improved visual homing. These results demonstrate that a simple error-driven mechanism can generate key features of learning walks under real-world conditions, supporting its plausibility as an explanation for active visual learning in desert ants. More broadly, this work illustrates how physical robots can be used to test biological models.

\section{Introduction}

To find their way home after foraging over long distances, desert ants combine path integration (PI) with learned visual information~\cite{Wehner2003,Collett2006}. PI provides a homeward vector, but its errors accumulate during travel, making visual memories important for locating the nest accurately~\cite{Merkle2006-hj,Huber2015-un}.  Ants acquire visual memories during learning walks (LWs), a series of exploratory trips undertaken before they begin foraging~\cite{Fleischmann2016-bd,Zeil2019}. With experience, ants generally walk faster, venture farther from the nest, follow straighter paths, and perform fewer scans. Their visual homing ability also improves in parallel~\cite{Fleischmann2016-bd,Deeti2021-hf}. Yet, alteration of the familiar panorama can cause learning behaviours to reoccur in experienced ants~\cite{Muller2010,Vega_Vermehren2020-lx}.

Snapshot-based navigation models explain how stored views can support visual homing~\cite{Cartwright1983,Zeil2003,Baddeley2012}. However, they do not explain when visual memories are acquired during LWs or how these walks are dynamically controlled. 
Recently, two computational models have been proposed to explain how LWs are dynamically controlled and how visual memories are acquired during these walks. 
\citet{Xiang2025} proposed a model in which panoramic images are encoded using rotation-invariant Zernike moments~\cite{Stone2018,Sun2020} and processed by a mushroom body model. The resulting familiarity signal controls the balance between exploration and homing during LWs, reproducing several characteristic features of these behaviours. In our previous work \cite{Misiek2026}, we proposed an alternative mechanism based on the angular mismatch between path integration (PI) and a learned visual compass (VC), reproducing complementary features such as changes in walking speed and pirouette frequency. Here, we test whether this mechanism can autonomously generate LWs on a physical robot.

In \citet{Misiek2026}, we proposed a different mechanism based on the angular discrepancy between the nest directions estimated by PI and a mushroom-body-inspired visual compass (VC). This error signal, $E_{\mathrm{PI/VC}}$, controls LW behaviour, while PI also teaches the VC to associate panoramic views with the nest direction, following \citet{Muller2010}. High error produces slow, spiralling walks with frequent pirouettes and early returns to the nest. As the VC learns, the error decreases and walks become longer and straighter. When the visual environment changes, PI and VC predictions diverge again, causing learning behaviours to reappear. In simulation, the model reproduced many features of desert-ant LWs. However, whether the same mechanism can generate LWs using real sensors and under real-world noise remains to be tested.

The transition from simulation to robotics is not trivial because simulations rely on assumptions that do not necessarily hold in the real world. This is a well-known problem in robotics, commonly referred to as the ``sim-to-real'' problem or ``reality gap''~\citep{Jakobi1995,Jakobi1997,Hofer2021, Webb2001}. [link to robotics] As such, manhy models have been tested on robots 

Insect-navigation models have already been tested on robotic platforms. Sahabot~\cite{Lambrinos2000} and AntBot~\cite{Dupeyroux2019} demonstrated ant-inspired PI using compass and odometry information. \citet{Stone2017-ub} implemented a central-complex-inspired PI circuit in embodied navigation, later transferred to a physical aerial robot by \citet{Stankiewicz2020}. 

Other robotic studies have investigated visual homing and route following. In particular, \citet{Gattaux2025} demonstrated outdoor visual homing using a mushroom-body-inspired model on the Antcar robot, with PI used during training to label panoramic views acquired during manually controlled learning walks. More recently, \citet{Ou2026BeeNav} used automated learning flights on a drone, during which PI trained a neural network to predict a home vector from omnidirectional images. Although these learning flights were automated, they did not reproduce the structure of bees' learning flights observed in experimental data. 
A key contribution of these robotic studies is that they confront insect-inspired navigation models with real-world sensory and motor noise, testing whether mechanisms developed in simulation remain effective under physical conditions. In this spirit, we test whether our previously proposed PI–VC mismatch mechanism can autonomously generate the characteristic structure and development of desert-ant learning walks on a physical robot.

Here, we reproduce the experimental arrangement of \citet{Fleischmann2016-bd}, using three cylinders placed in a grassy area on the University of Sussex campus. The model described and simulated in \citet{Misiek2026} is now implemented on a four-wheeled differential-drive robot equipped with a panoramic camera, an inertial sensor, wheel encoders, and GNSS tracking (see Methods and Fig.~\ref{fig:diagram_robot}). We test whether the model can support LWs and visual learning in the real world, with the robot autonomously deciding when to collect visual memories, scan, return to the nest, and transition to foraging. We show that the robot reproduces the main characteristics of desert-ant LWs, including expanding and straightening trajectories, decreasing pirouette frequency, and improved visual homing with experience. To our knowledge, this is the first robotic implementation of dynamically controlled learning walks. 


\begin{figure}
\begin{adjustwidth}{-2.2in}{0in}
\centering
\includegraphics[width=0.60\columnwidth,keepaspectratio]{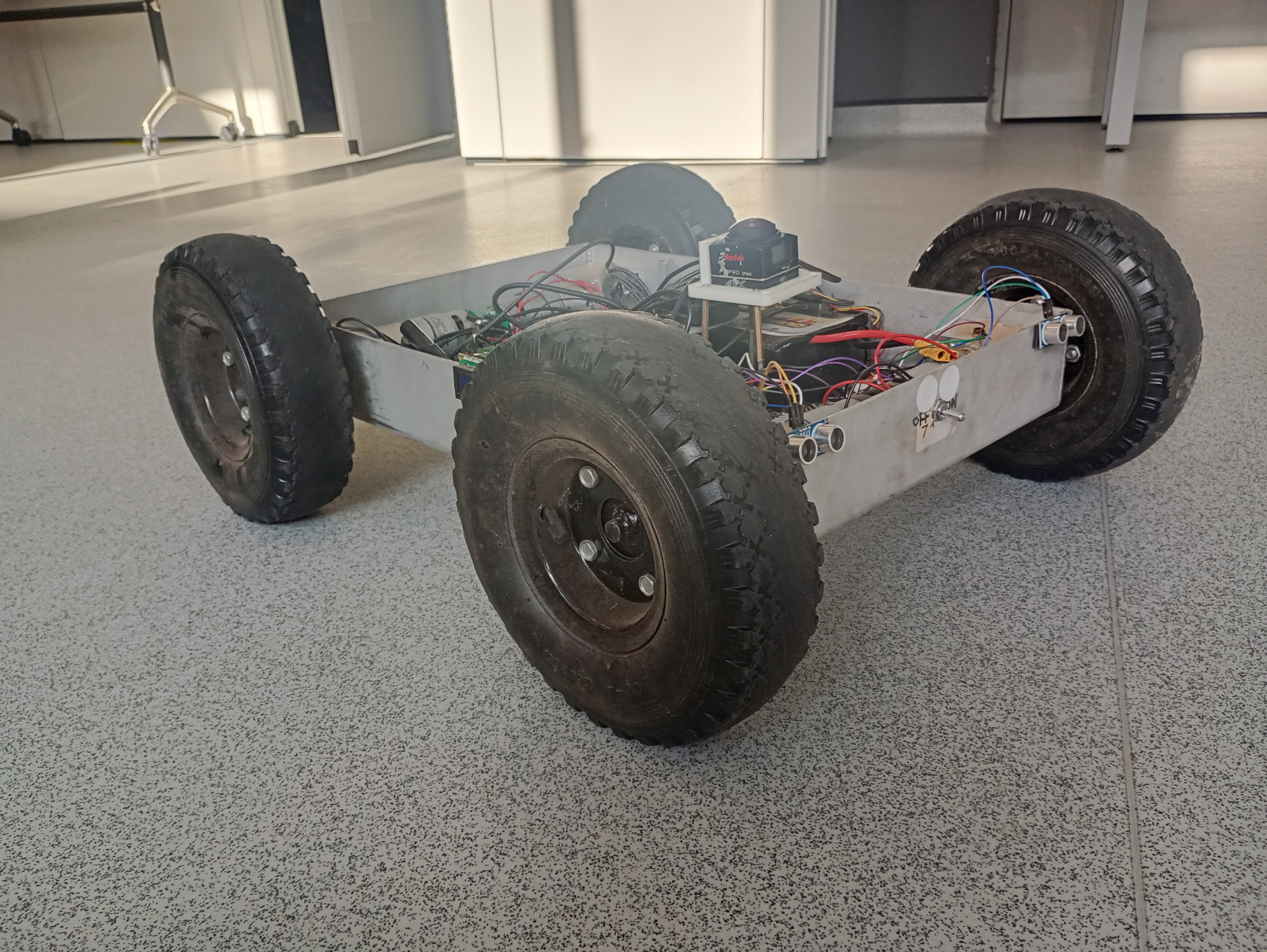}
\vspace{5mm}
\caption{
\textbf{Robot platform.}
Photograph of the custom-built robot used in the experiments. To give an indication of scale, the wheels are 27~cm in diameter and the top-mounted panoramic camera is positioned 29~cm above the ground. The two HC-SR04 ultrasonic distance sensors used for obstacle avoidance are visible at the front of the robot, on the right side of the image. See Section~\ref{sec:robot/methods/robot} for a full description of the platform.
}
\label{fig:robot}
\end{adjustwidth}
\end{figure}


\begin{figure}
    \begin{adjustwidth}{-2.2in}{0in}
    \centering    
    \includegraphics[width=15cm, keepaspectratio]{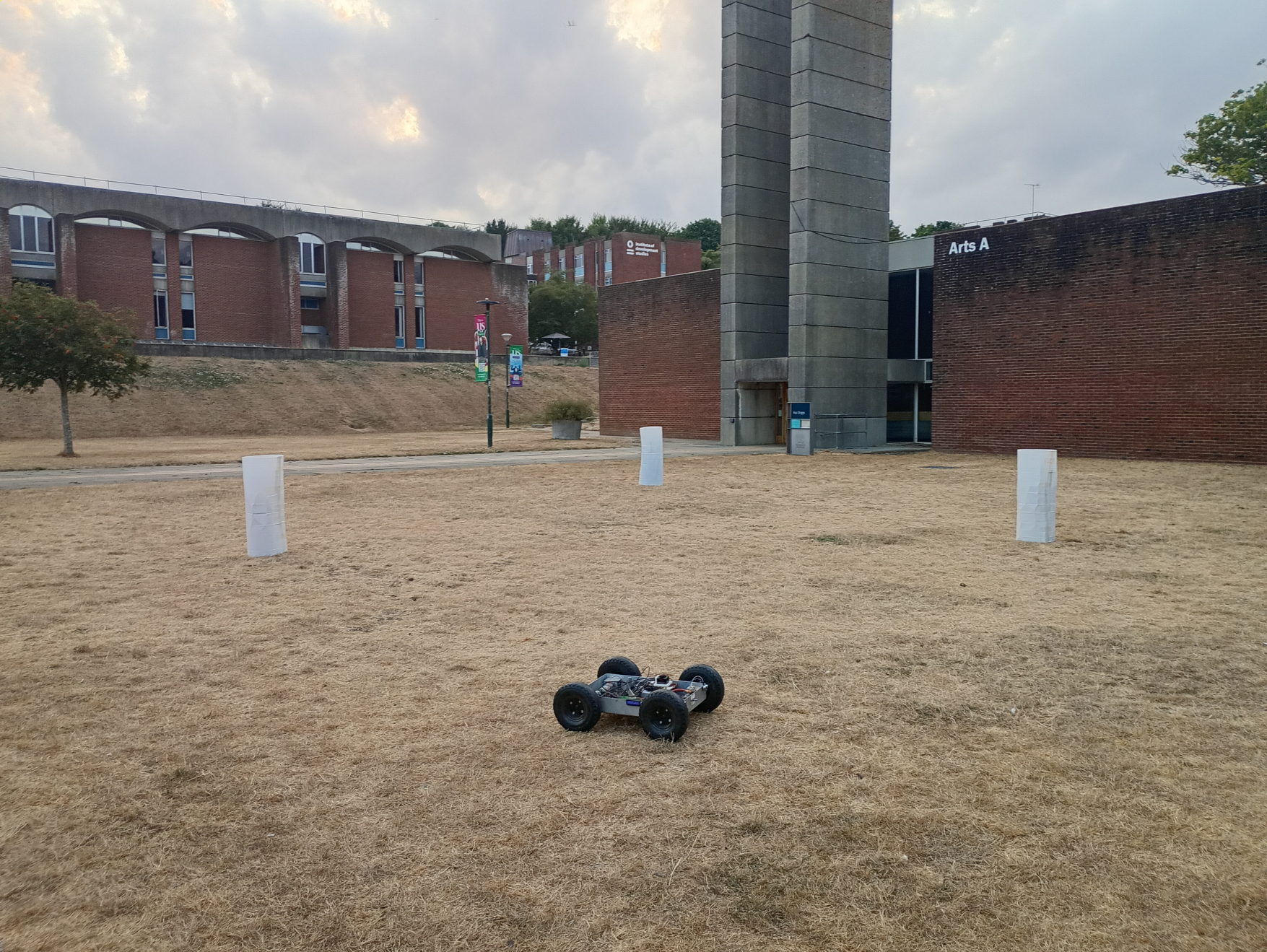}
    \vspace{5mm}
    \caption{
    \textbf{Experimental reproduction of \citet{Fleischmann2016-bd}.}
    Photograph of the robot in the foreground, with the three white cylinders reproducing a scaled-up version of the experimental layout of \citet{Fleischmann2016-bd}, near the Library Square at the University of Sussex. The sky was cloudy. The grass is unusually yellow due to the summer 2026 drought. The buildings in the background provided prominent visual landmarks compared with the largely featureless Tunisian desert environment used by \citet{Fleischmann2016-bd}.
    }
    \label{fig:robot_library}
    \end{adjustwidth}
\end{figure}

\section{Experimental protocol}
Before presenting the results, we first describe the experimental protocol here. The robotic platform and navigation algorithms are detailed in the Methods section at the end of the paper.

\subsection{Layout}
We reproduced the main experimental paradigm of \citet{Fleischmann2016-bd}  on a robotic platform (Fig.~\ref{fig:robot}). In the original experiment, three black cylinders 38~cm high and 22~cm in diameter were placed 2~m from the nest on a largely featureless Tunisian saltpan. Here, we doubled the scale of the original layout to accommodate the larger dimensions of the robot and GNSS positioning uncertainty, which was typically centimetre-scale (median horizontal accuracy 1.4~cm) but could occasionally deteriorate (Fig.~\ref{fig:robot_library}). Three white cardboard cylinders approximately 1~m high and 45~cm in diameter were therefore placed 4~m from the nest (Fig.~\ref{fig:robot_library}), and the 3~m threshold used in \citet{Misiek2026} to distinguish LWs from foraging walks and record pirouettes, was increased to 6~m. Because our experimental site contained a much richer visual panorama than the original saltpan, with prominent trees and buildings, we also took steps to improve the visibility of the artificial landmarks: the cylinders were made taller than a strict twofold scaling would require and white was chosen instead of black to increase the contrast against the dark green of the trees and the brownish-red and dark grey colors of the buildings (Fig.~\ref{fig:robot_library}).


\begin{figure}
    \begin{adjustwidth}{-2.2in}{0in}
    \centering    
    \includegraphics[width=15cm, keepaspectratio]{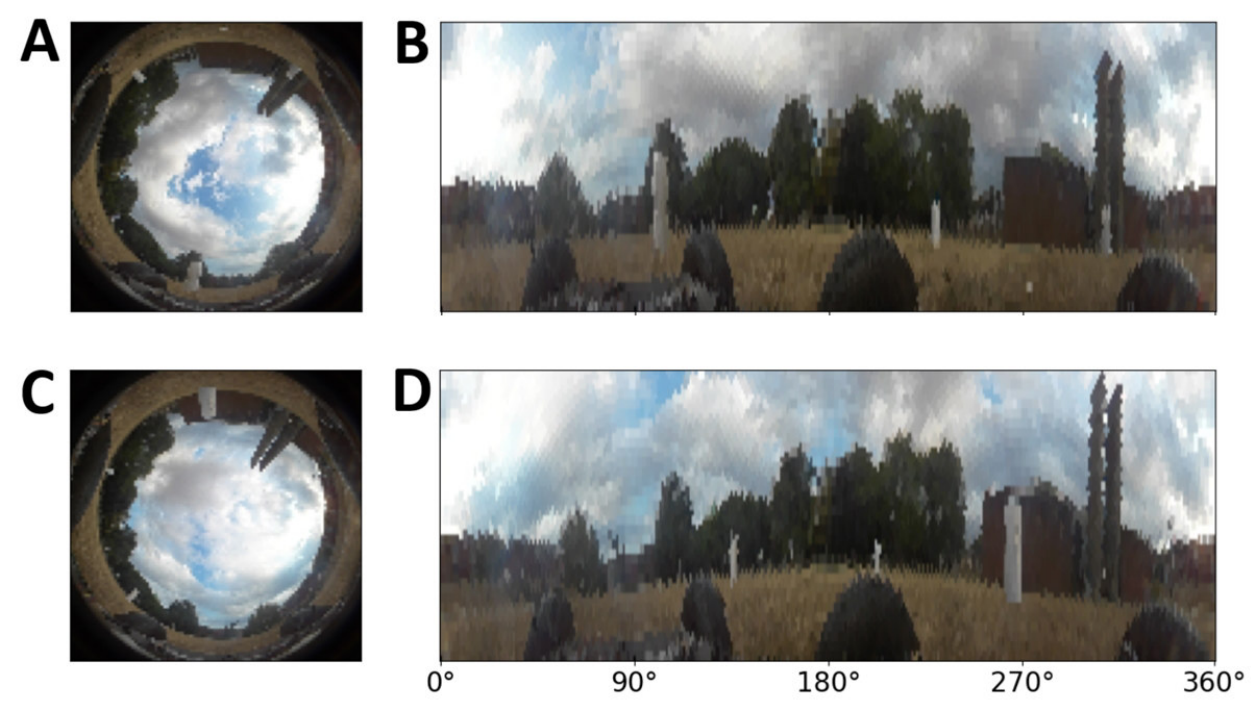}
    \caption{
    \textbf{Robot point of view through the Pixpro SP360.}
    \textbf{(A)} Raw 100$\times$100~px fisheye image.
    \textbf{(B)} Unwrapped panoramic view of the image in A. The upper 90$^\circ$, containing mostly sky, were excluded to reduce the influence of the sky and moving clouds on the visual-compass input. The three cylinders are visible but remain small relative to the surrounding landmarks such as trees and buildings.
    \textbf{(C)} Raw fisheye image acquired at the same orientation as A, but from a different position, twenty minutes after A. Distant background landmarks remain unchanged, whereas the nearby cylinders shift substantially. The clouds have also moved relative to A.
    \textbf{(D)} Unwrapped panoramic view of the image in C. Comparison of B and D also shows that the camera was not fully stabilized in pitch as the robot moved over uneven terrain. This can be seen as the yellow grassy ground is neither flat nor forming a similar horizon in B and D. Unfortunately, we did no record tilt and roll data (which the IMU had access to), so we could not plot camera pose data.
    }
    \label{fig:panoramic}
    \end{adjustwidth}
\end{figure}

Experiments were conducted on an approximately $30\times40~\mathrm{m}$ grassy area near the Library Square at the University of Sussex (Fig.~\ref{fig:robot_library}). A fixed nest position was defined near the centre of the experimental area, at $50.865538^\circ$N, $-0.08865920^\circ$W. The original and displaced nest positions, cylinder positions, and visual-homing release position were marked with wooden nails so that the experimental setup could be reproduced across the two days. Six independently trained agents were tested: four on 15 August 2026 between approximately 17:00 and 21:30, and two on 16 August 2026 between approximately 16:00 and 19:00. Both days were cloudy, and light rain interrupted some trials. The visual input was noisy, with moving clouds, frequent changes in illumination, and changes in camera pitch as the robot moved over uneven terrain (Fig.~\ref{fig:panoramic}).

\subsection{Protocol}

In the original experiment, \citet{Fleischmann2016-bd} followed individually marked newcomer ants over successive appearances outside the nest, recording their behaviour as they progressed from early LWs to foraging. At different stages of experience, ants were captured near the nest in a zero-vector state and transferred to a distant test field containing an identical landmark array. 
During visual homing tests, they were released 3~m from the fictive nest position and their search paths were recorded for five minutes. Because they were zero vector ants this test isolated homing based on learned visual cues, not PI.

We reproduce this test protocol with our robot: after each LW, the robot was displaced to a fixed release position 12~m southeast of the nest. This was farther than the 6~m expected from the approximately twofold scaling of the original protocol and was chosen to better demonstrate visual-homing performance. Visual homing was initiated by pressing a button on the joystick, after which the robot navigated autonomously for at least 90~s using only the visual compass, with no contribution from PI.

\paragraph{Here I summarize the experimental protocol:}

\begin{figure}
    \begin{adjustwidth}{-2.2in}{0in}
    \centering    \includegraphics[width=\columnwidth,height=.75\textheight,keepaspectratio]{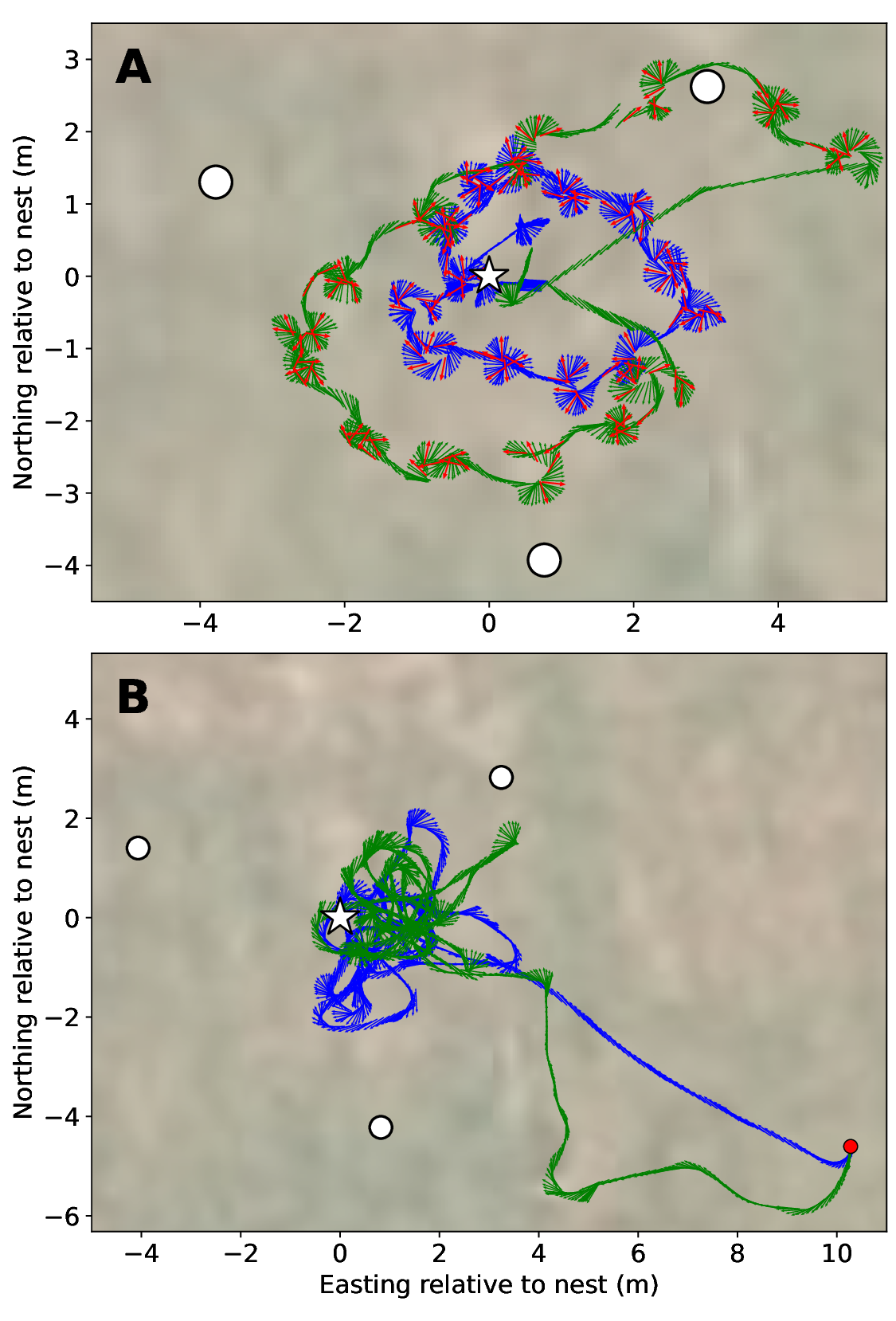}
    \caption{
    \textbf{(A)} Example of two consecutive LWs performed by a single agent. First LW in blue and second in green. Each arrow represents the robot's heading at 100~ms intervals. Red arrows indicate the robot's heading during snapshots. Nest position as a white star. Cylinders as white circles. The background is a 50\% whitened Google Earth image of the experimental site. 
    \textbf{(B)} Example of two consecutive ZV visual homing trajectories from an experienced agent. Seventh test in blue, eighth in green. Tests started 12 meters south-east of the nest (red dot mark). Both visual searches are centered roughly on the nest. Scale differs between A and B.
    }
    \label{fig:LW1_arrows}
    \end{adjustwidth}
\end{figure}

At the beginning of each experiment, I created a new agent by pressing a button on the control joystick (Fig.~\ref{fig:diagram_robot}). The robot was then placed at the marked nest position, its PI state was reset to ``zero'', and the first LW was started by pressing another button on the joystick. The agent navigated with the algorithm described in \citet{Misiek2026}. Briefly:

\begin{enumerate}
    \item The agent automatically selected a random allocentric heading from a uniform distribution between $0^\circ$ and $360^\circ$, rotated until it was within $20^\circ$ of that heading. This was implemented to promote a random and uniform exploration of all directions around the nest. From that point, the robot performed the LW autonomously using the algorithm introduced in \citet{Misiek2026}.

    \item PI--visual-compass mismatch controlled spiralling behaviour and the onset of pirouettes (see Methods).

    \item When the accumulated PI/VC error reached a threshold, the LW ended and the robot switched to PI homing until it returned to the estimated position of the nest and began searching.

    \item The robot was then displaced to the visual homing test, 12 meters southeast of the nest (see Fig.~\ref{fig:LW1_arrows}B and Fig.~\ref{fig:robot_library}).

    \item The robot performed visual homing for at least 90 seconds, using only the visual compass signal, and its search for the nest was recorded.

    \item I took the agent back to the exact position of the nest. Using the joystick, the PI vector was reset to ``zero'' and the visual memories from the previous LW were consolidated in long-term memory.
\end{enumerate}

\begin{figure}
    \begin{adjustwidth}{-2.2in}{0in}
    \centering
    \includegraphics[width=\columnwidth,height=.80\textheight,keepaspectratio]{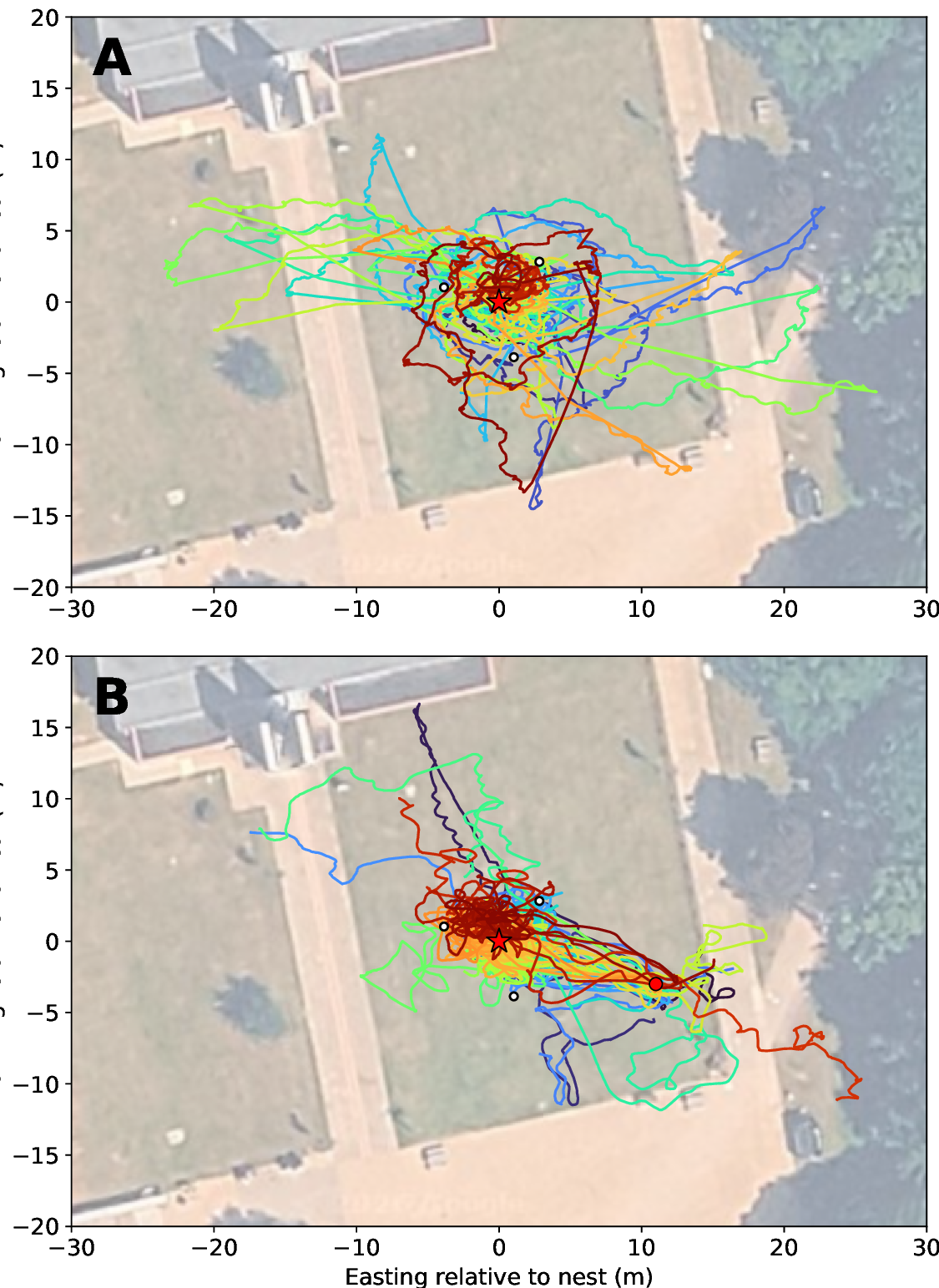}
    \vspace{5mm}
    \caption{
    \textbf{Top view of all 44 learning walks \textbf{(A)} and 54 zero-vector visual homing tests \textbf{(B)} conducted across the six trials.} The plotted area spans $60~\mathrm{m}$ in longitude and $40~\mathrm{m}$ in latitude, with the longest LW reaching 27.32~m from the nest. Colours are used only to distinguish the trajectories and have no additional meaning.
    }
    \label{fig:all_trajectories}
    \end{adjustwidth}
\end{figure}

All agents completed this cycle -- a LW followed by an immediate visual homing test -- at least seven times. We aimed to complete nine LWs per agent, but some sequences were shortened by rain or unexpected Jetson restarts. Unlike the original experiment, in which each ant was tested only once and then discarded, the same agent was tested after each successive LW. 

\paragraph{Shifted cylinders experiment}
With agents 3 and 4, after nine LWs, the cylinders were displaced 4~m northwest and an additional visual homing test was conducted. A final test was then performed with the cylinders removed. These manipulations tested whether visual homing was controlled predominantly by the cylinders relative to other landmarks such as trees and buildings, which were not available in the original experiment.

All data were recorded using GNSS with RTK corrections. See Section~\ref{sec:GNSS_reliability} for an accuracy report for each experiment.

\section{Results}

Across the six experimental sequences, 44 LWs and 54 visual-homing tests were recorded (Fig.~\ref{fig:all_trajectories}A,B). The robot traveled a total of 5151.02~m over 2~h 48~min 57.56~s of recordings.

\subsection{Learning-walk behaviour changed with experience}

\begin{figure}
    \begin{adjustwidth}{-2.2in}{0in}
    \centering    \includegraphics[width=\columnwidth,height=1.0\textheight,keepaspectratio]{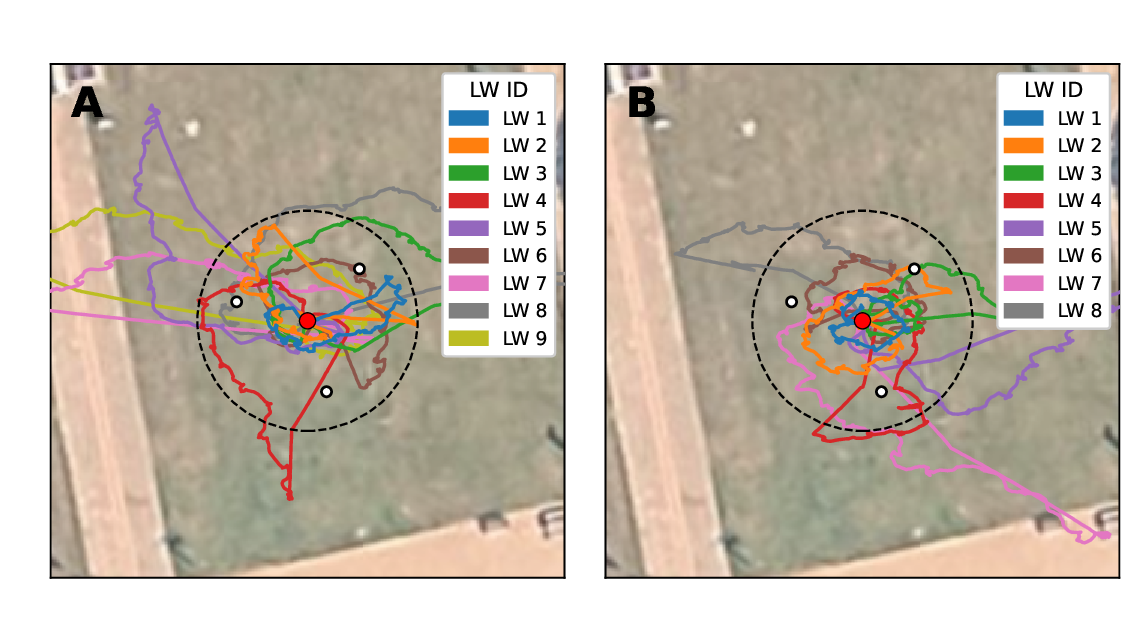}
    \caption{
    \textbf{All consecutive LWs of two agents.}
    \textbf{(A)} All consecutive LWs for agent 2. Eight walks were recorded, each shown in a different color (blue for the first walk, orange for the second, etc.). The 6~m radius around the nest is indicated by a grey dashed circle. The first two LWs remain close to the nest. Subsequent walks are more directed and extend beyond the 6~m radius, and are therefore labelled as foraging walks. The plotted area is a $24 \times 24~\mathrm{m}$ square. \textbf{(B)} Another example of a series of consecutive LWs, for agent 4. Nine walks were recorded. Cylinders as white dots.}
    \label{fig:full_training}
    \end{adjustwidth}
\end{figure}

\begin{figure}
    \begin{adjustwidth}{-2.2in}{0in}
    \centering
    \includegraphics[width=\columnwidth,height=.95\textheight,keepaspectratio]{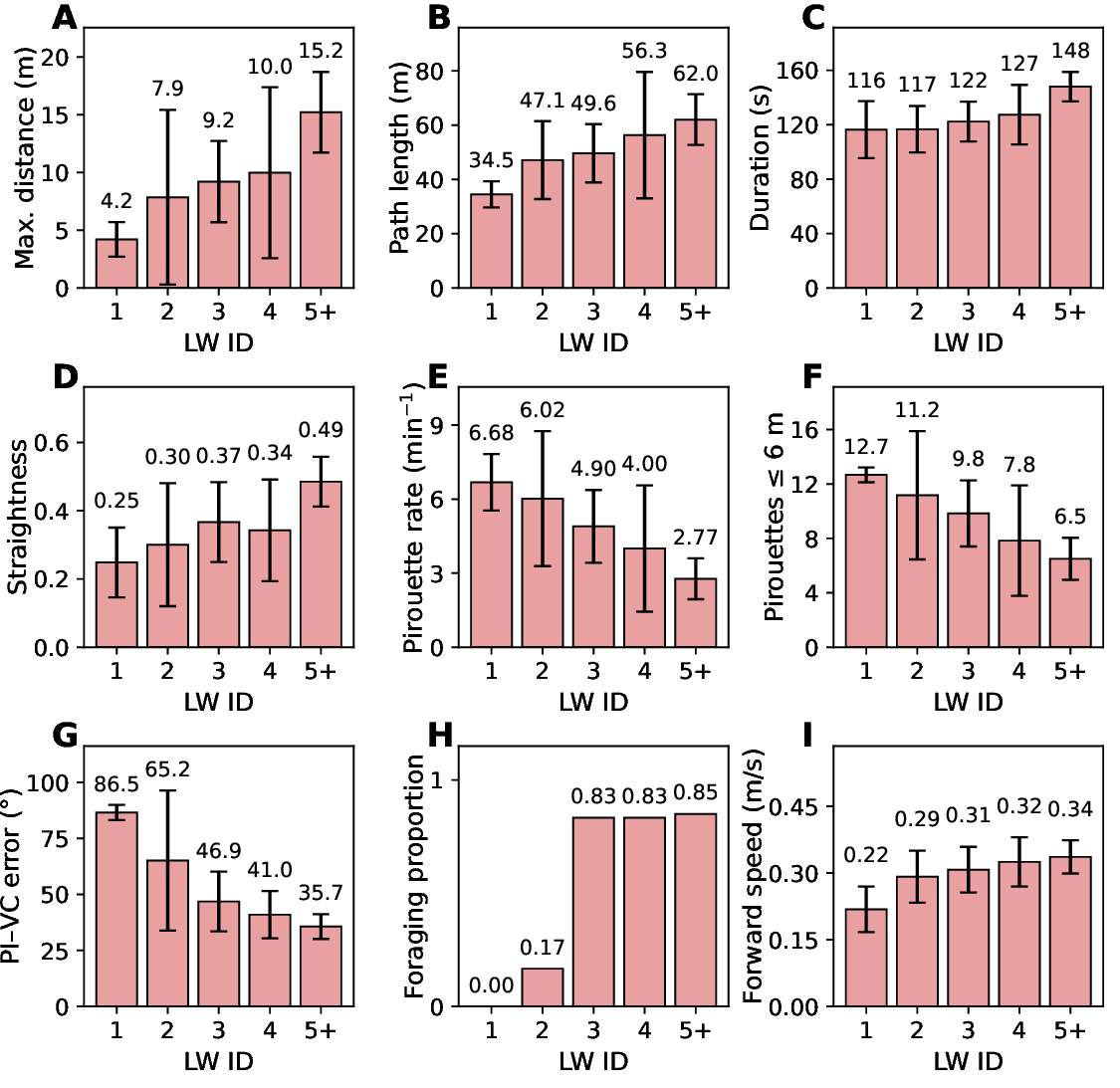}
    \vspace{5mm}
    \caption{
    \textbf{Changes in LW behaviour with increasing experience.}
    Measurements are shown for the first four LWs, with all subsequent walks grouped as 5+.
    \textbf{(A)} Maximum distance from the nest increased approximately threefold between the first LW and the 5+ group.
    \textbf{(B)} Total distance travelled increased across LWs.
    \textbf{(C)} LW duration remained broadly similar, with a slight increase across walks.
    \textbf{(D)} Path straightness doubled between the first LW and the 5+ group.
    \textbf{(E)} Pirouette frequency decreased two-fold.
    \textbf{(F)} Number of pirouettes performed within a 6~m radius of the nest, following the same criterion \cite{Muller2010}, decreased twofold.
    \textbf{(G)} PI--VC angular error, which control learning behaviours, decreased across LWs
    \textbf{(H)} Proportion of foraging walks ($>6$~m from the nest), increased and reached a plateau by the third LW.
    \textbf{(I)} Forward driving speed of the robot during successive LWs in meters per seconds.
    }
    \label{fig:results}
    \end{adjustwidth}
\end{figure}

Across the six agents, the robot progressively transitioned from short LWs, characterised by spiralling around the nest, to longer, more directed trajectories (Fig.~\ref{fig:all_trajectories}A). This progression is evident across the complete LW sequences of agents 2 and 4 (Fig.~\ref{fig:full_training}A,B). The expansion of the LWs is also visible across an example of two consecutive walks from the same agent (Fig.~\ref{fig:LW1_arrows}A), shown at 10~Hz (100~ms intervals), with the pirouettes and snapshots clearly identifiable. 

Because performances generally plateaued from approximately the fifth LW onward, LWs 1--4 were analysed separately and subsequent LWs were pooled into a 5+ category (Fig.~\ref{fig:results}). For each measure, a linear slope was fitted across these five groups within each of the six agents, and the resulting slopes were tested against zero using a two-sided one-sample $t$-test with Benjamini--Hochberg FDR correction across behavioural measures.

Measures of spatial expansion increased consistently with experience. Maximum distance from the nest increased from 4.21~m in LW1 to 15.22~m in the 5+ group (Fig.~\ref{fig:results}A; mean within-agent slope $=2.45~\mathrm{m}$ per category, $t_5=16.19$, $p_{\mathrm{FDR}}<0.001$), while total travelled distance increased from 34.49 to 62.02~m (Fig.~\ref{fig:results}B; $6.88~\mathrm{m}$ per category, $t_5=3.62$, $p_{\mathrm{FDR}}=0.017$). Across LWs, the maximum distance reached from the nest was 27.32~m, while total distance travelled ranged from 9.53~m in the shortest LW to 93.74~m in the longest (which was classified as a foraging walk as it went farther than 6~m from the nest).

The proportion of walks extending beyond 6~m similarly increased from 0 in LW1 to 0.85 in the 5+ group (Fig.~\ref{fig:results}H; $+23.5$ percentage points per category, $t_5=7.59$, $p_{\mathrm{FDR}}=0.002$). The first ``foraging'' walk occurred after one to three LWs, with one agent leaving the 6~m radius on the second walk, four agent on the third walk and the remaining agent on its fourth walk. 

LW duration increased from 116.4~s in LW1 to 148.0~s in the 5+ group (Fig.~\ref{fig:results}C; $8.44~\mathrm{s}$ per LW, $t_5=2.76$, $p_{\mathrm{FDR}}=0.040$). Over the same period, straightness increased from 0.248 to 0.485 (Fig.~\ref{fig:results}D; $0.053$ per LW, $t_5=4.44$, $p_{\mathrm{FDR}}=0.009$). LW duration ranged from 94.3 (classified as LW) to 191.4~s (classified as foraging walk, >6m from nest).

Simultaneously, scanning behaviour decreased. Pirouette frequency fell from $6.68$ to $2.77~\mathrm{min}^{-1}$ between LW1 and the 5+ group (Fig.~\ref{fig:results}E; slope $=-1.01~\mathrm{min}^{-1}$ per category, $t_5=-6.98$, $p_{\mathrm{FDR}}=0.002$). We performed a similar analysis to \citet{Muller2010}, who recorded the number of pirouettes in a 3~m radius of the nest. We doubled that radius in our analysis. The number of pirouettes performed within 6~m of the nest decreased from 12.7 to 6.5 (Fig.~\ref{fig:results}F; $-1.57$ pirouettes per category, $t_5=-8.26$, $p_{\mathrm{FDR}}=0.002$). 

Forward speed also increased with experience (Fig.~\ref{fig:results}I), with mean speeds of 0.218, 0.292, 0.307, 0.325, and 0.336~m/s for LW1, LW2, LW3, LW4, and the 5+ group, respectively (mean within-agent slope $=0.0269~\mathrm{m/s}$ per category, $t_5=4.56$, $p_{\mathrm{FDR}}=0.009$). The speed distribution was wide and the robot reached maximum speeds of up to 1.25~m/s (5~km/h) during LWs.

\begin{figure}
    \begin{adjustwidth}{-2.2in}{0in}
    \centering
    \includegraphics[width=\columnwidth,height=1.\textheight,keepaspectratio]{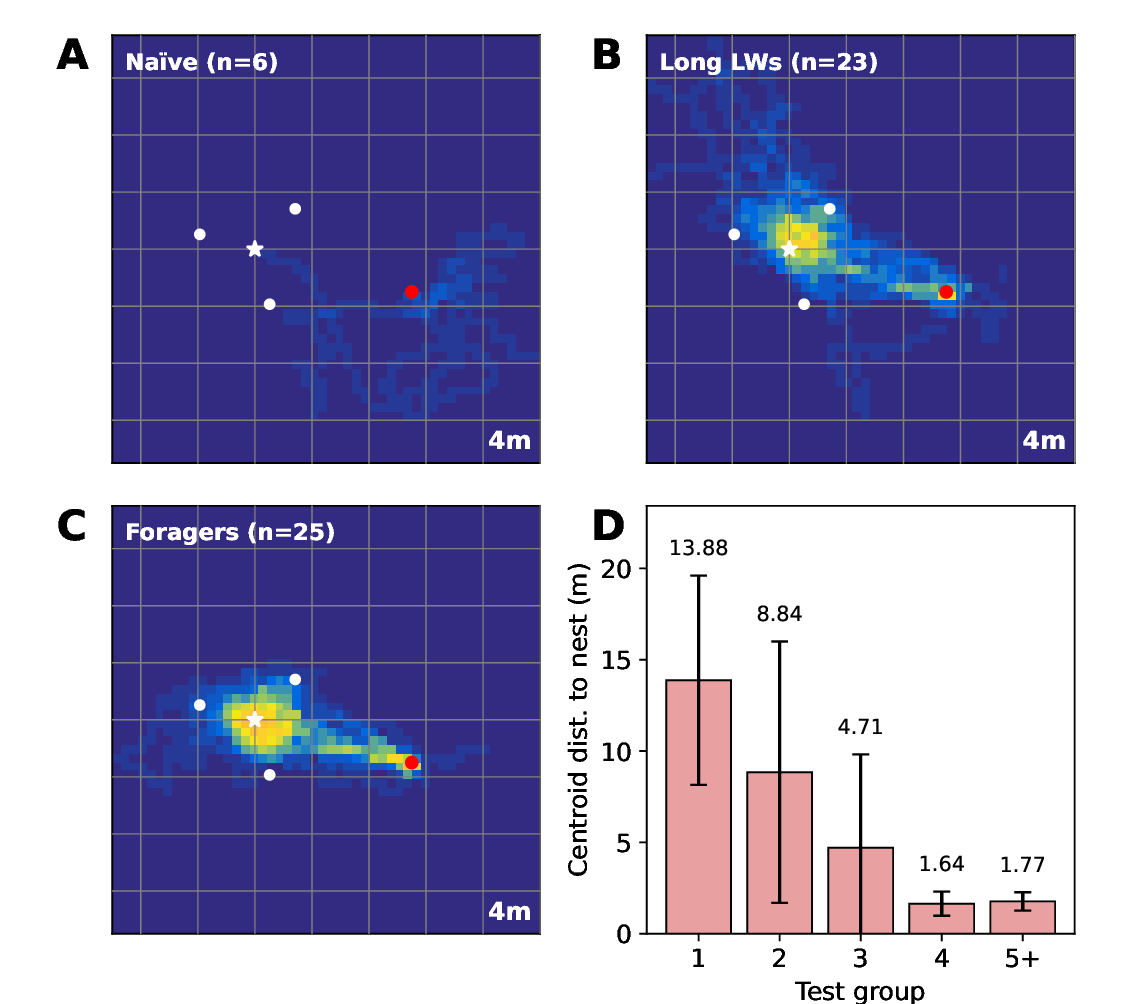}
    \vspace{5mm}
    \caption{
    \textbf{Visual homing tests performance at different levels of experience.}
    \textbf{(A)} Heatmap of all naive test trajectories. Cylinders as white dots, nest position as a white start. Grid size is 4~m by cell. Test start position as a red dot. ($n = 6$)
    \textbf{(B)} Heatmap of all homing tests from long-LW agents. Searches remain spread out, but are clearly centred on the nest. ($n = 23$)
    \textbf{(C)} Heatmap of all tests from foraging agents ($n = 25$). Trajectories are more directed and searches are more tightly centred on the nest. Long LWs data not shown. 
    \textbf{(D)} Centroid distance to the nest during visual homing tests as a function of test ID. Search centroids were farther from the nest during early tests and became progressively closer after successive LWs, showing improved homing performance with experience.
    }
    \label{fig:zero_vector_visual_homing_tests}
    \end{adjustwidth}
\end{figure}

\begin{figure}
    \begin{adjustwidth}{-2.2in}{0in}
    \centering \includegraphics[width=\columnwidth,height=.95\textheight,keepaspectratio]{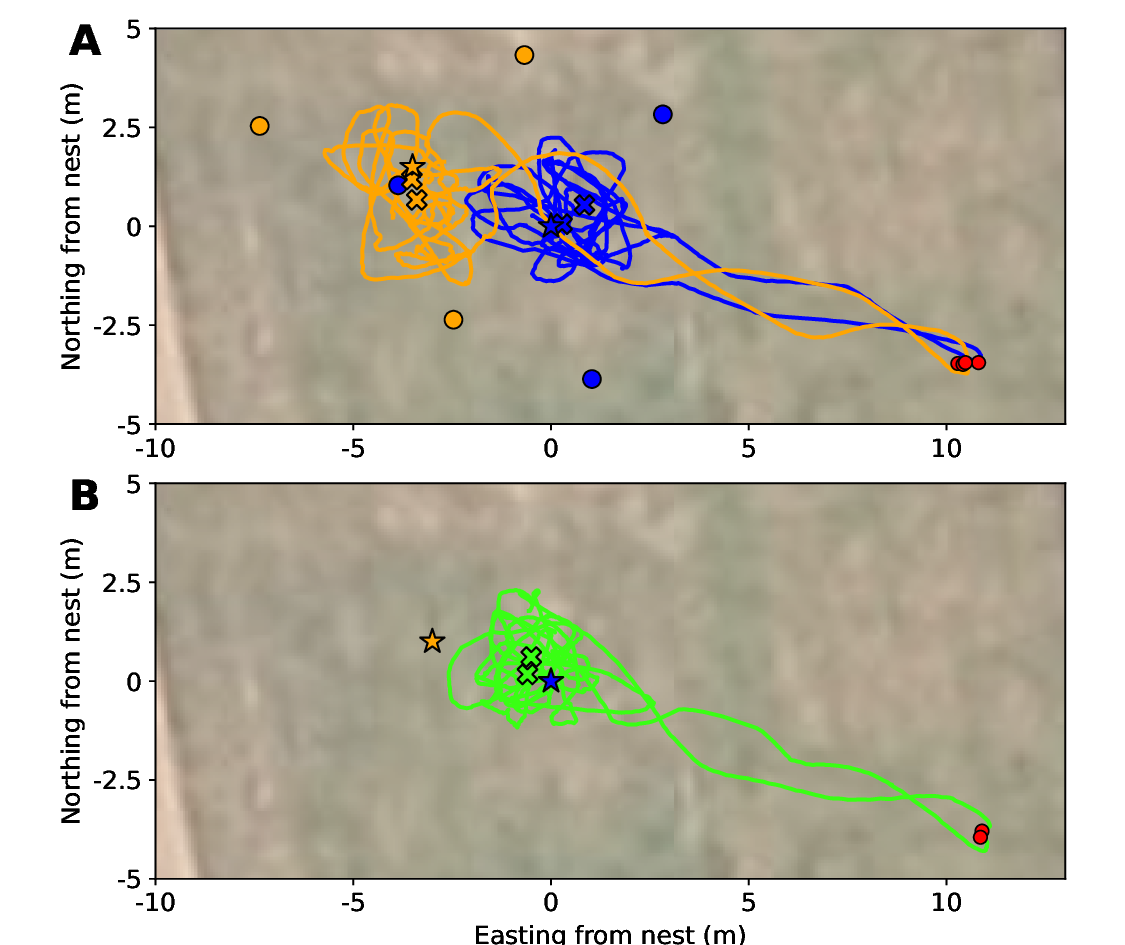}
    \vspace{5mm}
    \caption{
    \textbf{Influence of the cylinders}
    \textbf{(A)} Search behaviour was affected by the position of the cylinders. The agent was trained using cylinders positions shown in blue. Control tests shown in blue. All tests were conducted with an experienced agent which performed seven LWs prior to the tests. The orange trajectory shows the search after the cylinders were shifted 4~m to the northwest. The orange, shifted configuration was used only during tests. The search centre shifted in the same direction as the cylinders, showing that they strongly influenced visual homing. This is reminiscent of the classic landmark-displacement experiment of \citet{Tinbergen1932-ex}, in which shifting landmarks around a wasp nest displaced the subsequent search.
    \textbf{(B)} Nevertheless, the cylinders were not the only landmarks available to the agent. When they were removed, the agent was still able to perform a tight search centred near the true nest position, indicating that other visual landmarks, such as trees and buildings, could also support homing.
    }
    \label{fig:test_shift_cylinders}
    \end{adjustwidth}
\end{figure}

PI--visual-compass angular error decreased from $86.5^\circ$ to $35.7^\circ$  between LW1 and the 5+ group (Fig.~\ref{fig:results}G; $-10.29^\circ$ per LW, $t_5=-6.83$, $p_{\mathrm{FDR}}=0.003$), showing increasing agreement between the two navigation systems with accumulating experience, and explaining the overall transition from learning oriented behaviours to more directed walks. 
We also analysed the accuracy of the raw visual compass signal and the temporally filtered VC signal from our CX model in non naive agents (excluding the first LW). The mean absolute angular error between PI and VC was 51.00$^\circ$ for the raw signal and 36.92$^\circ$ after CX integration. Between consecutive predictions only 100~ms apart, the mean angular distance between consecutive predictions (recorded 100~ms apart in the CSV) fell from 44.90$^\circ$ to 14.09$^\circ$, while the mean circular SD of ten successive predictions within a 1-s windows decreased from 51.29$^\circ$ to 20.90$^\circ$. The raw VC signal was therefore very noisy, particularly given the 44.90$^\circ$ change over only 100~ms, and temporal integration in the CX ring model stabilized it.

\begin{figure}
    \begin{adjustwidth}{-2.2in}{0in}
    \centering    \includegraphics[width=\columnwidth,height=.95\textheight,keepaspectratio]{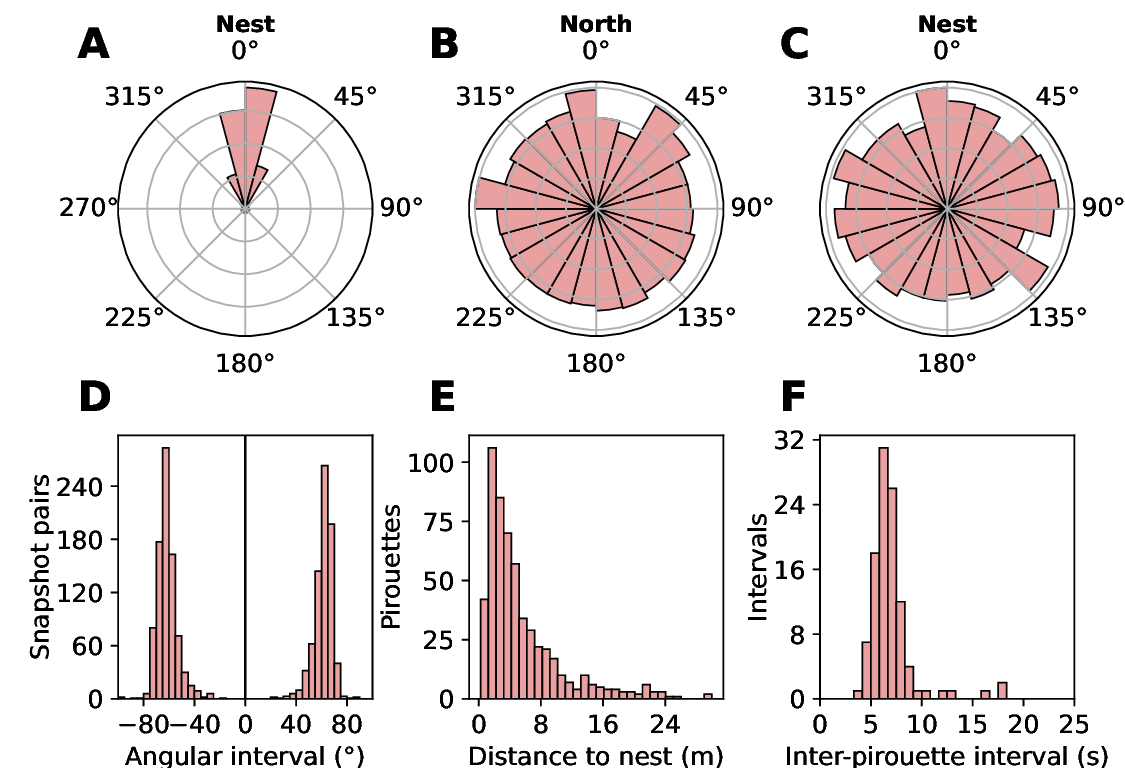}
    \vspace{5mm}
    \caption{
    \textbf{Pirouette and snapshot statistics.}
    \textbf{(A)} PI angular error relative to the ground-truth nest direction provided by GNSS.
    \textbf{(B)} Distribution of snapshot headings relative to North, showing an approximately uniform distribution.
    \textbf{(C)} Distribution of snapshot headings relative to the nest direction, also approximately uniform.
    \textbf{(D)} Angular interval between consecutive snapshots, with two peaks around $-65^\circ$ and $+65^\circ$.
    \textbf{(E)} Distance of snapshots from the nest. Most snapshots, and therefore most pirouettes, occurred within 5~m of the nest.
    \textbf{(F)} Inter-pirouette time interval averaged across LW1 and LW2, with a peak around 7~s.
    }
    \label{fig:pirouettes_snapshots}
    \end{adjustwidth}
\end{figure}

\subsection{Visual homing performances improved with experience}
Across the 54 zero-vector tests (Fig.~\ref{fig:all_trajectories}B), visual homing improved with experience. Tests were grouped as naive (no previous LW), short-LW (<1.4m from nest), long-LW (less than two walk >6m from nest) and foragers (at least 2 previous walk > 6~m), following the terminology used by \citet{Fleischmann2016-bd} (see methods). No agent fitted the short-LW category at any test in our data. Naive agents searches were not centered on the nest (Fig.~\ref{fig:zero_vector_visual_homing_tests}A). Searches after long-LW experience were clearly centred on the nest, while those of experienced ``forager'' agents were even more tightly clustered around the nest position (Fig.~\ref{fig:zero_vector_visual_homing_tests}B,C).

The improvement in homing performance with experience was particularly clear across successive tests: search-centroid distance to the nest decreased from 13.93~m during the first test walk to 1.81~m in the 5+ group (Fig.~\ref{fig:zero_vector_visual_homing_tests}D). This was also evident within individual sequences: the seventh and eighth ZV tests of one experienced agent both converged toward the nest area from the release location (Fig.~\ref{fig:LW1_arrows}B).

\subsection{The cylinders influenced visual homing but were not the only useful landmarks}
In the original experiment of \citet{Fleischmann2016-bd}, the ants inhabited a nearly featureless saltpan, with the array of three artificial cylinders providing the only prominent visual cues. There is no featureless desert near the University of Sussex so our experiment was conducted in a much richer environment. Whether the array of cylinders were relevant cues at all, compared to larger landmarks like trees and buildings, is therefore an important question.
We tested the influence of the array by moving the three cylinders 4~m northwest. This shifted the centre of the visual search in the same direction (Fig.~\ref{fig:test_shift_cylinders}A), showing not only that the cylinders contributed to visual homing, but that they were the strongest visual cue for the visual compass model, despite appearing small relative to the other landmarks in the surrounding panorama (Fig.~\ref{fig:panoramic}B,D).
Finally, we conducted an additional test in which the three cylinders were removed to determine whether they were the only cues used by the robot. Despite their removal, the robot still performed a compact search near the original nest position (Fig.~\ref{fig:test_shift_cylinders}B), indicating that other local cues such as trees and building, not present in the original experiment, also supported homing.

\subsection{Pirouettes and snapshot acquisition}

\begin{figure}
    \begin{adjustwidth}{-2.2in}{0in}
    \centering    \includegraphics[width=\columnwidth,height=.95\textheight,keepaspectratio]{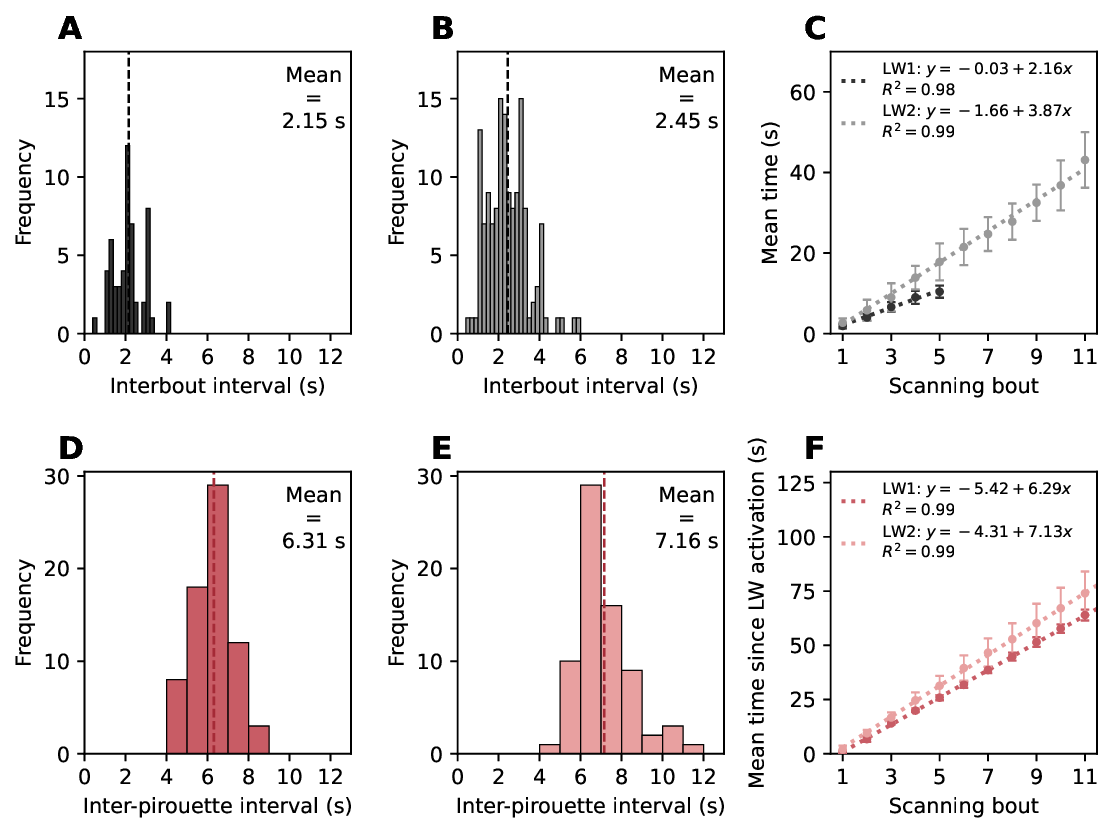}
    \vspace{5mm}
    \caption{
    \textbf{Comparison of scanning-bout timing in ants and robotic learning walks.}
    \textbf{(A, B)} Distribution of interbout intervals during the first (A) and second (B) learning walks of ants, digitized from \citet{Deeti2021-hf}, with mean intervals of approximately 2.15 and 2.45~s, respectively.
    \textbf{(C)} Mean time of successive scanning bouts in ants, with linear regression slopes of 2.16 and 3.87~s/bout for LW1 and LW2, respectively ($R^2=0.98$ and $0.99$).
    \textbf{(D, E)} Distribution of inter-pirouette intervals during the first (D) and second (E) learning walks of six robotic agents, with mean intervals of $6.31 \pm 1.01$~s ($n=70$) and $7.16 \pm 1.35$~s ($n=71$), respectively (mean $\pm$ SD).
    \textbf{(F)} Mean time of successive robotic pirouettes, with regression slopes of 6.29 and 7.13~s/bout for LW1 and LW2, respectively ($R^2=0.9998$ and $0.9995$).
    Error bars in (C) and (F) represent SD. Data in (A--C) were approximately digitized from \citet{Deeti2021-hf}.
    }
    \label{fig:interval_snapshot}
    \end{adjustwidth}
\end{figure}

The mean absolute PI angular error relative to the true GNSS nest direction was $1.42^\circ$ Fig.~\ref{fig:pirouettes_snapshots}A). The corresponding circular distribution had a mean resultant length of $r=0.943$ and a circular SD of $18.55^\circ$ ($n=377$).
Averaged over the six agents, snapshot headings were approximately uniformly distributed relative to both North (Fig.~\ref{fig:pirouettes_snapshots}B) and the nest direction (Fig.~\ref{fig:pirouettes_snapshots}C). Thus, views were sampled in all directions, promoting a relatively uniform distribution of learning across the eight directionally tuned MBONs. Individual agent data also showed a uniform distribution. The total number of pirouettes recorded within the 6~m radius was 377, with a minimum of 40, a maximum of 84, and a mean of 63 pirouettes per agent. During pirouettes, the recorded mean absolute angular speed was $64.16^\circ/\mathrm{s}$, and a pirouette completion took 4.39 seconds on average.
The mean number of snapshots per pirouette was 5.20, with a mean SD of 0.85 and a mean interquartile range of 0.67. 

\begin{figure}
\begin{adjustwidth}{-2.2in}{0in}
\centering
\includegraphics[width=\columnwidth,height=.95\textheight,keepaspectratio]{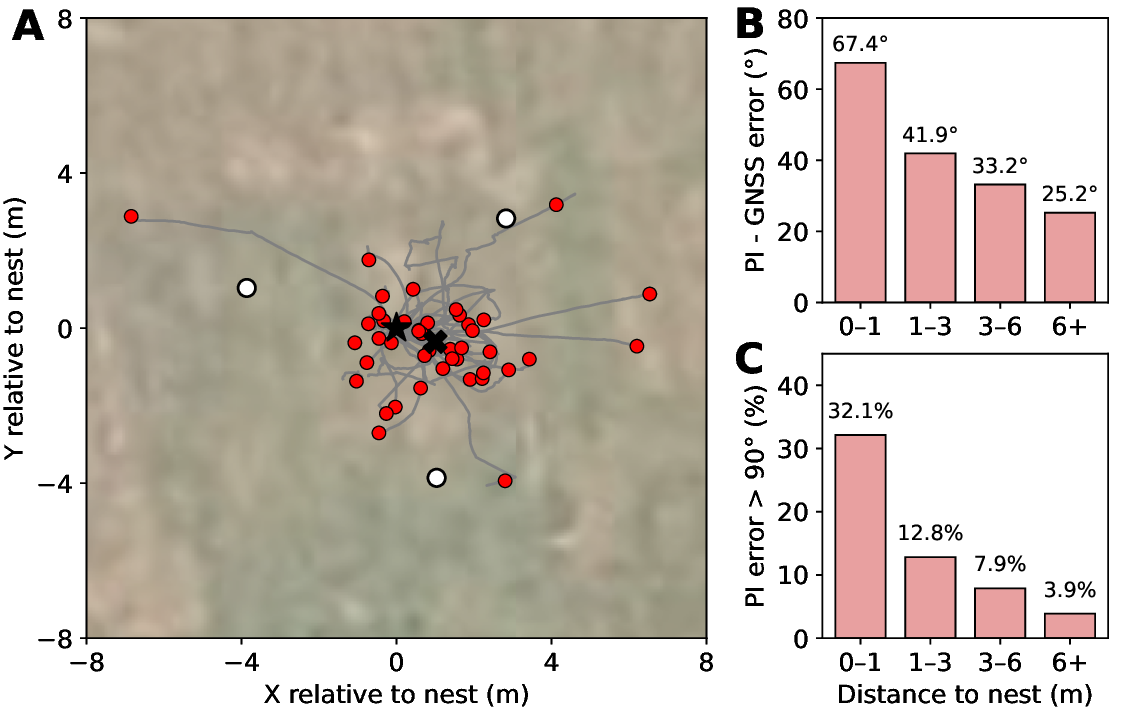}
\vspace{5mm}
\caption{
\textbf{PI search error after LWs.}
\textbf{(A)} PI search locations after the 44 LWs. Red dots indicate the positions at which the PI search started, the black star indicates the true nest position, and the black cross indicates the centroid of the search locations. Grey lines show the trajectories produced when the robot was manually driven from the PI search position back to the nest. The background is a 50\% whitened Google Earth image of the experimental site. The plotted area is $16\times16$~m.
\textbf{(B)} Percentage of snapshots for which the angular error between the PI-estimated nest direction and the GNSS-derived ground-truth exceeded $90^\circ$, for four distance bins from the nest. 
\textbf{(C)} Absolute angular error between the PI-estimated and GNSS-derived nest directions for the same distance bins. Only measurements with GNSS horizontal accuracy below 10~cm were included in B and C. Average angular error decreased with higher distance to the nest.
}
\label{fig:PI_error}
\end{adjustwidth}
\end{figure}

Angular distance between consecutive snapshots peaked around $-65^\circ$ and $+65^\circ$ (Fig.~\ref{fig:pirouettes_snapshots}D).
Most snapshots, and therefore most pirouettes, occurred within 5~m of the nest~(Fig.~\ref{fig:pirouettes_snapshots}E).

\subsection{Timing of pirouettes during learning walks}

\citet{Deeti2021-hf} investigated the timing of scanning bouts during the first two LWs of \textit{Melophorus bagoti}. A scanning bout was defined as a sequence of body rotations performed at one location, interrupted by stopping phases. From video recordings of eight first and six second learning walks, they found that the timing of successive scanning bouts was approximately linear, with regression slopes of 2.16~s/bout for LW1 ($R^2=0.98$) and 3.87~s/bout for LW2 ($R^2=0.99$). The interbout intervals were variable, both between and within individuals. Digitizing their published distributions gave approximate mean intervals of 2.15~s for LW1 and 2.45~s for LW2 (Fig.~\ref{fig:interval_snapshot}A--C).

We analysed the timing of pirouettes during the first two LWs of our six agents. The minimum observed interval was approximately 4~s. During LW1, the agents performed 76 pirouettes, producing 70 inter-pirouette intervals, with a mean of $6.31 \pm 1.01$~s (mean $\pm$ SD). During LW2, they performed 77 pirouettes, producing 71 intervals, with a mean of $7.16 \pm 1.35$~s (Fig.~\ref{fig:interval_snapshot}D,E). The timing of successive pirouettes was highly linear, with regression slopes of 6.29~s/bout for LW1 ($R^2=0.9998$) and 7.13~s/bout for LW2 ($R^2=0.9995$; Fig.~\ref{fig:interval_snapshot}F).

Overall, the robotic agents performed pirouettes at longer intervals than the ants observed by \citet{Deeti2021-hf}, but showed similar temporal patterns: relatively regular intervals, a minimum delay between successive bouts, a roughly bell-shaped distribution of intervals, and an increase in the mean interval from LW1 to LW2.

\subsection{PI search error during LWs}

Across the 44 PI returns, the mean distance between the PI search location and the actual nest location was 2.1~m (Fig.~\ref{fig:PI_error}A). The average length of the LWs was 51.8~m, with the smallest LW being 27.4~m, and the longest 92.5~m. The maximum PI search error was 8~m. The centroid of the 44 PI search locations was displaced by 1.06~m from the true nest position, indicating a bias --which origin is unknown--, toward the South-East direction. This displacement was greater than expected from sampling variation under a zero-centred Gaussian distribution with the same spatial covariance as the experimental data (parametric Monte-Carlo test, $p=0.001$, $n=44$). 

PI angular error during LWs was measured at snapshot locations using the absolute angular difference between the PI-estimated nest direction and the reference nest direction obtained from the GNSS signal. Snapshot measurements were grouped according to distance from the nest (0-1~m, 1-3~m, 3-6~m, and $>6$~m), and only measurements with GNSS horizontal accuracy below 10~cm were included. PI angular error depended strongly on distance from the nest (Fig.~\ref{fig:PI_error}B,C). Within 0-1~m, the mean absolute error was $67.39^\circ$, with 32.1\% of snapshots exceeding $90^\circ$. PI error to the true nest direction decreased to $41.93^\circ$ at 1-3~m, with 12.8\% of snapshots exceeding $90^\circ$, and to $33.18^\circ$ at 3-6~m, with 7.9\% snapshots exceeding $90^\circ$. Beyond 6~m, the mean absolute error was $25.24^\circ$, and only 3.9\% of snapshots exceeded $90^\circ$. Snapshots acquired close to the nest were therefore substantially more likely to be associated with large PI angular errors, and the PI teaching signal could even act counter productively by  associating views to MBONs encoding for the opposite direction of the nest. 

\subsection{Summary}
In this section, we showed that the main features of desert ant LWs were reproduced in the robot: walks progressively expanded and became straighter, scanning behaviour decreased, and overall, LWs eventually transitioned into foraging-like walks. In parallel, visual-homing performance improved with accumulating experience.

\section{Conclusion}


\paragraph{Our active learning model replicates LW behaviour and support learning on a robotic platform, under real-world conditions}

The main result of this work is that the active-learning model developed in \citet{Misiek2026} also operates on a physical robot. PI-visual-compass mismatch was sufficient to regulate the transition from short, pirouette-rich LWs to longer and straighter trajectories and replicate known features of desert ants Lws \cite{Fleischmann2016-bd, Muller2010, Zeil2019}. As experience accumulated, PI-visual-compass disagreement decreased, pirouette frequency decreased, walks expanded, and visual homing improved. To the best of my knowledge, this is the first robotic implementation reproducing the progression from the first LWs to the first foraging-like walks in ants. This shows that the model is not restricted to idealized simulations, but can operate in a real sensorimotor loop with noisy visual input, imperfect PI, wheel slippage, uneven terrain, and changing weather. Unfortunately, while we were able to quantify PI error, we could not analyze camera stabilization data as it was absent from the recording process at the time of the experiment.

\paragraph{PI search error in our robot reached similar levels as what is reported in the desert ant LW and insect PI literature}

The mean absolute PI angular error during pirouettes was $1.42^\circ$, with $r=0.943$ and a circular SD of $18.55^\circ$ ($n=377$; Fig.~\ref{fig:pirouettes_snapshots}A). This spread was comparable to that of nest-oriented fixations during ant LWs: SD=$23.5^\circ$ ($r=0.916$, $n=163$) in \textit{Ocymyrmex robustior}~\cite{Muller2010} and SD=$16.0^\circ$ ($r=0.962$, $n=11$) in \textit{Cataglyphis aenescens}~\cite{Fleischmann2017}. The robot's PI search error was smaller, but in the same order of magnitude as reported in the PI literature. The robot searches had an average error of 2.1~m after LWs of a mean length of 51.8~m, compared with a median error of 5.9~m reported in \textit{Cataglyphis fortis} after a 50~m path~\cite{Huber2015-un}.

\paragraph{PI provided a noisy but sufficient teaching signal to scaffold visual learning}
PI accuracy is particularly important in our active-learning model because PI provides both the homing signal and the teaching direction used to train the visual compass during LWs. Close to the nest, the short PI vector produces large angular uncertainty, meaning that some LWs are associated with bad or even counterproductive snapshots, which encode an imprecise or even the opposite direction of the nest. Nevertheless, the distribution remains centred on the true nest direction, so these errors tend to cancel across the many snapshots acquired over successive LWs. Thus, despite substantial noise, especially close to the nest, PI still provided a useful teaching signal on average.

\paragraph{Sources of PI error}

In \citet{Misiek2026}, idealised simulations included independent, fixed step-wise compass noise, following \citet{Wystrach2015}. Here, no artificial noise was added to PI, so errors arose from either the terrain (i.e. slope, uneven ground with holes and branches) or the robot itself, including sensor noise, and systematic biases such as (slightly) different wheel diameters or persistent wheel slip. Unlike independent step-wise noise, which can partly cancel over time, sustained biases can accumulate rapidly. In our experiments, the PI signal showed a statistically significant systematic bias, with the estimated nest position shifted by 1.06~m towards the southeast. We included several measures to avoid biases arising from the robot itself, such as randomising spiralling and scan direction, which should reduce systematic effects caused by factors such as unequal wheel diameters, to which encoder-based PI is particularly sensitive. The remaining bias therefore likely originated from the terrain.

Wheel slippage was particularly frequent during pirouettes and turning (personal observation), a known source of dead-reckoning error in skid-steer robots~\cite{Mandow2007,JingangYi2009}. A substantial fraction of PI error may therefore have accumulated during pirouettes, although this remains to be tested rigorously. During rotation, opposite encoder signals from the left and right wheels cancelled out, but not when one wheel slipped. Thus, ignoring translational encoder updates during pirouettes could potential help suppressing pirouette induced PI error. The size and the nature of the body of the navigational agent may expose them to very different sources of noise, and pirouettes induced error is likely to be far less consequential in ants. Indeed, the robot was approximately 50~cm wide, so each wheel travelled about 1.6~m during a full $360^\circ$ pirouette, providing substantial opportunity for slippage and accumulating errors. In ants, by contrast, the corresponding rotational displacement is only on the millimetre scale and is dwarfed by the distance travelled between successive pirouettes. Additionally, the probability of an ant's legs slipping during rotation is likely different from that of a skid-steer robot. Nevertheless, despite these morphological differences, our results demonstrate that the model can operate under real-world conditions, including sources of sensory and motor noise that are difficult to anticipate or reproduce in simulation.

Although our robotic implementation introduces realistic sensory and mechanical sources of PI error, it still omits a potentially important biological limitation: the finite representational capacity of the neural circuit. In our model, the home vector is distributed across eight directional units, but their activity has no upper bound. In contrast, the CPU4 model of \citet{Stone2017-ub} uses bounded neural activity, such that sufficiently long paths eventually saturate the PI memory. A biological PI system may therefore lose precision as the encoded vector grows because of finite neural resolution and dynamic range, independently of accumulated sensory noise or systematic bias. Such representational limitations could potentially constitute a major source of PI error in insects. More generally, constraints including finite neuron numbers, firing ranges, spiking variability and memory decay are absent from our implementation. Interestingly, behavioural work in \textit{Cataglyphis fortis} found that PI uncertainty increased with home-vector length rather than with the total distance walked~\cite{Merkle2010}. This is consistent with a limitation associated with representing increasingly large home vectors, although it does not prove that finite neural capacity is the underlying cause.

\paragraph{Robustness to visual noise}

The experiments show that the model can operate despite real-world constraints and noise, such as changes in illumination and weather, PI error, camera pitch, wheel slippage, and uneven terrain. However, each learning sequence was completed within a single experimental session, never lasting more than an hour. Even over such a short period, the panorama changed substantially, particularly through changes in cloud cover and illumination, and the model continued to perform well without explicit sky segmentation, which has been implemented in previous insect navigation models using natural images \cite{Basten2010,Stone2014,Stone2016,Differt2015, Differt2016}. The only step we took to reduce the influence of the sky on the visual compass was to remove a 90$^\circ$ cone from the dorsal region of the panoramic visual input, which never contained any landmarks. However, a substantial portion of the sky remained accessible to the visual compass \ref{fig:panoramic}. With the present model, a realistic navigation task could be conducted by distributing training over the 2--4 day timescale over which ants perform LWs~\cite{Fleischmann2016-bd,Fleischmann2017}. The robot could then perform repeated long outward journeys followed by homing over subsequent days, mimicking ant foraging trips under different weather conditions and at different times of day. This would test whether homing remains robust across the temporal and visual variability experienced by foraging ants, while also providing the opportunity to assess the benefit of sky segmentation in a realistic navigation task, if two experimental groups were tested, one with sky segmentation and one without.

\paragraph{Our pirouette model is a useful simplification of the ants scans during LWs}

Unlike our simulated model~\cite{Misiek2026}, which included both full and partial pirouettes with stochastic snapshot acquisition, the robot performed only full $360^\circ$ pirouettes, acquiring snapshots at a fixed frequency during rotation, with variability in angular intervals arising from motor noise.
The angular distance between successive fixations of the same scan followed a similar bimodal distribution to that reported by \citet{Freas2026} in \textit{Melophorus bagoti}. Like ants, our model produced equal proportions of right- and left-directed turns, as expected from the underlying Bernoulli process. A similar process is implemented in the model to randomly choose between clockwise and anti-clowise spiralling at the beginning of a LW. Randomized scans direction prevented the introduction of systematic biases in the PI and visual-compass systems and promoted more uniform sampling of the panorama. Nevertheless, the distribution reported by \citet{Freas2026} is averaged across ants, and whether consistent preferences for scan direction occur at the individual level remains to be tested. In our robot, the angular distance between consecutive snapshots peaked around $-65^\circ$ and $+65^\circ$ (Fig.~\ref{fig:pirouettes_snapshots}D). \citet{Freas2026} reports almost twice smaller saccadic interval, with their reported distribution exhibiting peaks at -$\sim35^\circ$ and $\sim35^\circ$.
On average, our robot took 4.39 seconds to complete a pirouette, \citet{Deeti2021-hf} reported a median pirouette duration of approximately 0.8~s in \textit{Melophorus bagoti}, with individual bouts ranging from about 0.3 to 1.7~s, while two examples of pirouettes shown by \citet{Fleischmann2017} in \textit{Cataglyphis} lasted approximately 1.5 and 2~s.
We also find that the mean number of snapshots per pirouette was 5.20, with a mean interquartile range of 0.67. This is comparable to the median of four stopping phases per pirouette reported in \textit{Cataglyphis noda}, although with a larger interquartile range of 1.75~\cite{Grob2017}. The lower variability in our robot stems from the simple pirouette mechanism, with invariable $360^\circ$ rotations, with snapshots acquired at a stable 1.3~Hz rate, whereas \textit{C.~noda} performs more complex scans comprising both full and partial pirouettes~\cite{Fleischmann2017} and also possibly frequent reversal of scan direction as observed in \textit{Melophorus Bagoti} \cite{Freas2026}. 
Finally, the robot captured snapshots while rotating during pirouettes, without the fixations observed in ants \cite{Fleischmann2017, Freas2026}, and this did not seem to prevent learning (see methods). Overall, our simplistic robotic pirouette model is closer to the uninterrupted, $360^\circ$ voltes observed in \textit{Cataglyphis} ants \cite{Fleischmann2017}, than pirouettes. 
Ant scans therefore appear to be faster, more complex, typically spanning a smaller amplitude than our robot’s $360^\circ$ pirouettes, and also contain more densely spaced fixations. Nevertheless, our simple pirouette model fulfilled its role in supporting visual learning and promoting uniform sampling of the panorama.

\paragraph{Our pirouette-generation process reproduces the inter-scan interval distribution of \textit{Melophorus bagoti}}

Unlike our simulated model~\cite{Misiek2026}, which triggered pirouettes through Bernoulli trials, producing approximately exponential inter-pirouette intervals, the robot used an error-accumulation mechanism, producing more regularly spaced pirouettes with a bell-shaped interval distribution and an apparent "refractory period".
In our robot, the average inter-pirouette interval during the first and second LW showed a peaked distribution, with a minimum observed interval of 4~s, and a mean of 6.54~s during LW1 and 7.96~s during LW2 (Fig.~\ref{fig:pirouettes_snapshots}F). A similar peaked distribution of inter-scanning-bout intervals was reported by \citet{Deeti2021-hf} during the first LWs of \textit{Melophorus bagoti}, although scanning occurred considerably more frequently, with mean intervals of 2.16~s during LW1 and 3.87~s during LW2. Very short intervals were also rare in the ant data, suggesting an apparent ''refractory period'' of approximately 1~s, compared with 4~s in our robot. In our robot, this bell shaped distribution and refractory period arises from the error-accumulation process described in Section~\ref{sec:robot/methods/robot}. The model therefore reproduced three temporal features of ant scanning: a peaked interval distribution, an apparent refractory period, and an increase in inter-scan intervals from LW1 to LW2. This suggests that a similar error accumulation process could control the onset of scans during ants LWs, spacing scans more regularly than a memoryless random process. 
By accumulating PI--VC angular error rather than, for example, time or distance, the model ensures that pirouette behaviour disappears once the visual compass learns and PI and VC strategies start to agree. In contrast to LWs, \citet{Deeti2023} found that inter-scanning-bout intervals during homing followed an approximately exponential distribution, consistent with a random-rate Poisson process, suggesting that scan triggering may differ between learning and homing.

\paragraph{Limitations of our replication: size of the robot}
We used a 15~kg, approximately 50~cm-wide robot to model navigation in desert ants that are roughly 1~cm long and weigh only a few milligrams~\cite{Dupeyroux2019,McMeeking2012}. Using large robots to model small insects is common in robotic implementations of insect navigation: Sahabot measured $42\times42\times28$~cm and weighed approximately 8~kg~\cite{Lambrinos2000}, AntBot had a diameter of 45~cm and weighed 2.3~kg~\cite{Dupeyroux2019}. The QCar used by \citet{jesusanmi_investigating_2024} measured approximately $39\times21\times21$~cm and weighed 2.7~kg. Finally, Antcar used by \citet{Gattaux2025} measured approximately $24\times20\times13$~cm. Nevertheless, recent work has moved towards considerably greater miniaturisation, with for example AntFlie weighing only 33~g and spanning approximately 9~cm~\cite{Gattaux2026}. The large size and weight of the robot were necessary to accommodate the onboard hardware, provide enough battery capacity for several hours of uninterrupted sequences of LWs and tests, traverse rough outdoor terrain without being repeatedly stopped by small obstacles and needing human intervention, and reach sufficient speeds for long-range navigation over hundreds of metres.

\paragraph{Limitations of our replication: scaled up dimensions}

We doubled the dimensions of the experiment because we initially expected substantially larger GNSS errors than the mean error of 1.40~cm ultimately obtained, and anticipated that increasing the spatial dimensions would reduce the relative impact of this positioning error. Additionally, being half a meter in size, the robot itself was larger than many early ants LWs, which often remain within 30~cm of the nest \cite{Zeil2019, Jayatilaka2018-ns}. This provides an additional justification for scaling up the spatial dimensions of the experiment. Accordingly, cylinders were placed 4 rather than 2~m from the nest, their diameter increased from 22 to 45~cm, and the LW threshold was increased from 3 to 6~m. We argue that the larger scale of our protocol does not affect the scope of our results. The relevant visual features were scaled accordingly (see Results), and the robot itself was already many times larger than an ant, making an exact reproduction of the protocol scale not necessarily meaningful. The parameters of the active-learning model were thus adjusted so that the robot generated (twice) larger versions of the LWs observed in ants. Nevertheless, the robot did not reproduce the very short LWs, extending only tenths of centimeters from the nest, reported in ants~\cite{Fleischmann2017,Zeil2019,Jayatilaka2018-ns}; indeed, none of our trajectories would fall within the ``short LW'' (<0.70~cm from nest, 140~cm in our work) category described by \citet{Fleischmann2017}. This indicates that the model parameters and steering mechanisms could still be refined to better match ants' behaviours. 
Having a lower viewpoints or sufficient ventral field of view to sample nearby visual cues may also help close range navigation. Our robot camera, by contrast, was mounted 29~cm above the ground, and its FOV extended only $27.5^\circ$ below the horizon, thus increasing the influence of distant cues in the visual input. Future experiments aiming to reproduce very short LWs at the true scale of LWs may therefore benefit from smaller, potentially legged robotic platforms.

\paragraph{Limitations of our replication: richer panorama}

Visual-homing abilities improved with experience: experienced agents produced directed trajectories and compact searches around the nest, whereas naive agents did not. Displacing the three cylinders shifted the centre of the visual search in the same direction, showing that they strongly influenced the visual compass, while their removal did not prevent homing, showing that the robot had also learned information from the wider panorama. This richer panorama differed from the original experiment of \citet{Fleischmann2016-bd}, in which the artificial cylinders were the only available landmarks, and may have made the environment easier to learn. 

It is possible that this difference influenced the transition to foraging-like walks which occurred after only one to three LWs, earlier than the 2--7 LWs reported by \citet{Fleischmann2016-bd}. 
Alternatively, the apparently early transition to longer walks may simply reflect that the motor sequence promoted relatively large LWs from the outset, even in naive agents. Indeed, we have seen that no LW were classified as `short LWs'' according to the terminology of \citet{Fleischmann2016-bd}, which indicates that our robot's early LWs were simply larger than ants LWs, and that the foraging threshold may therefore have been crossed earlier than real ants did. 
Finally, the earlier transition may be because the parameters selected in our model shifted the exploration--exploitation trade-off towards exploitation. For example, spiraling was triggered when the instantaneous PI--VC angular error exceeded $25^\circ$; lowering this threshold would likely have prolonged LWs before the robot escaped the 6~m nest radius. This would shift the exploration--exploitation trade-off towards greater investment in learning around the nest before initiating foraging. 


\paragraph{Limitations of the model: when to stop learning walks}

In \citet{Misiek2026}, we argued that the active-learning algorithm requires a way to stop learning when the visual environment is uninformative or cannot be learned. The present experiments highlight a related limitation: the model also needs to determine when to stop expressing LW behaviour altogether. Desert ants perform a limited series of nest-centred LWs, spanning from centimetres to a few metres, before switching to much longer foraging journeys extending tens or even hundreds of metres from the nest~\cite{Zeil2019,Buehlmann2012b,Huber2015-un}. In contrast, our model has no such transition: the longest of the 54 recorded walks reached 27~m from the nest despite being preceded by eight LWs. A more complete model should therefore suppress the LW-specific behavioural rules beyond the nest vicinity: pirouettes and nest-centred spiralling should cease, and PI--VC error should no longer accumulate towards a threshold that triggers premature homing. In particular, a large PI--VC discrepancy far from the nest should not cause the agent to abandon a foraging journey and return home. Importantly, ending LW behaviour does not imply ending visual learning. Ants continue to acquire visual information along their foraging routes, and recent behavioural work suggests that route views are acquired through behaviours distinct from pre-foraging nest-centred learning walks~\cite{Freas2025}. Recent robotic models similarly distinguish nest-centred visual homing from route following, using different organisations of visual memory for these two navigational regimes~\cite{Gattaux2025,Gattaux2025Route}.

\paragraph{Toward a fully autonomous learning-walk sequence}

After each PI-guided return, the robot had to be manually repositioned at the nest and its PI vector reset before the next LW could begin. PI was sufficient to return the robot to the vicinity of the nest, but not to localise the nest precisely, and without manual correction PI error would have accumulated across successive LWs. Achieving a fully autonomous LW sequence therefore requires the robot first to localise and reach the nest, and then reliably detect that it has actually arrived, allowing the PI state to be reset before the next LW. This distinction is important in ants: familiar cues at the nest entrance are not sufficient to reset PI; reset occurs after entering the nest~\cite{Knaden2006resetPI}. 
A visual solution to nest detection was recently demonstrated by \citet{Gattaux2025}, who used a dedicated nest MBON to modulate robot speed according to the familiarity of nest-centred views. Across three trials, the robot stopped 0, 0.02 and 1.04~m from the nest. Although promising, such variability would be problematic for our application, as resetting PI before the robot has actually reached the nest (1.04~m away) would introduce an error into the subsequent LW. A fully autonomous sequence will therefore require a clear and reliable signal that the nest has actually been reached before accumulated PI error can be reset between successive LWs.

\section{Methods}

\subsection{Robotic platform}
\label{sec:robot/methods/robot}

\begin{figure}[p]
    \begin{adjustwidth}{-2.2in}{0in}
    \centering
    \includegraphics[width=15cm, keepaspectratio]{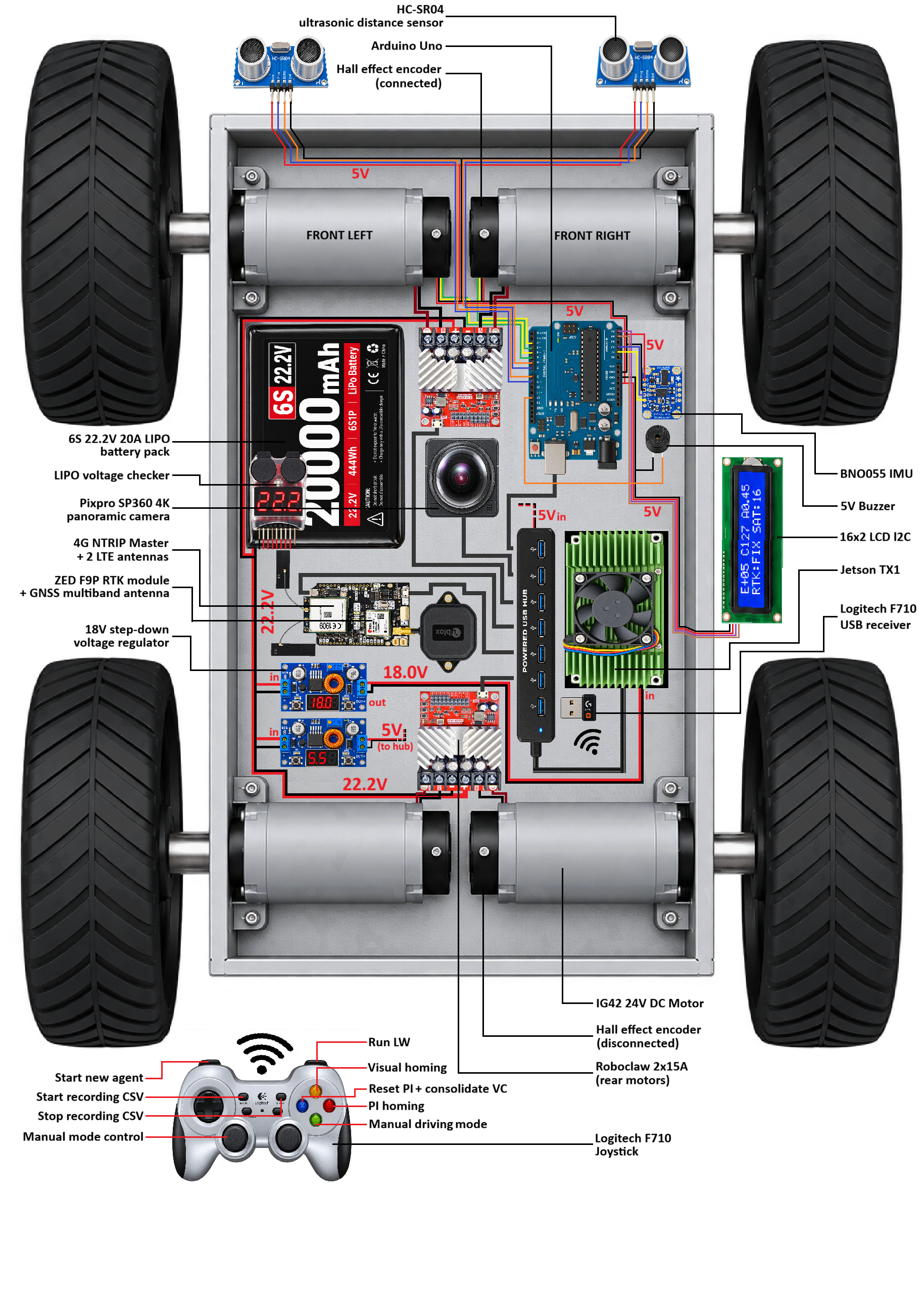}
    \captionsetup{justification=centering}
    \caption{Schema of the robot platform}
    \label{fig:diagram_robot}
    \end{adjustwidth}
\end{figure}

Experiments were conducted on a custom-built terrestrial robot (Figure~\ref{fig:robot}). High-level control, navigation algorithms and visual processing were executed on an NVIDIA Jetson TX1, while low-level sensor acquisition, including heading, wheel encoders, and obstacle detection, was handled by an Arduino Uno. Motor control was performed by two RoboClaw 2$\times$15A motor controllers. The robot chassis was 82~mm high, 383~mm wide, and 558~mm long, and the camera was mounted 290~mm above the ground. The wheels had a diameter of 270~mm. The robot weighed 15~kg. At full motor power, it could reach a forward speed of 1.7~m/s ($\sim$6~km/h).

The robot was powered by a 6S 22.2~V, 20000~mah LiPo battery.
The battery supplied the motor controllers directly, while voltage regulators provided 18~V to the Jetson TX1 and 5~V to a powered USB hub used to connect the camera, 2 motor controllers, joystick, Arduino and 2 GNSS ports to the Jetson (Figure~\ref{fig:diagram_robot}). The Jetson could be accessed locally by connecting a screen or remotely via SSH. The main Python control program was configured to launch automatically at startup, and experimental data were recorded onboard as CSV files, one file for each LW or test. The robot could be manually controlled during experiments using a Logitech F710 joystick. 

Motor actuation was achieved using four IG42 DC gear motors driving four pneumatic wheels independently. The robot used differential drive, with the left and right wheels driven at different speeds to control turning. The wheels were partially deflated to increase traction and provide a degree of passive suspension, reducing vibrations and wheel slip on uneven terrain. Each motor was equipped with an integrated Hall-effect quadrature encoder (Figure~\ref{fig:diagram_robot}). During the experiments, the two front wheel encoders were connected and their signals were combined in a signed signal to estimate forward and backward motion for PI. Rotational motion caused the integrated signal between the two wheels to cancel out. One of the rear Hall-effect encoders was not functional, which is why only the front encoders were used to compute the PI signal. PI was still surprisingly accurate with this setup. 

Heading information was obtained from a Bosch BNO055 inertial measurement unit (Figure~\ref{fig:diagram_robot}). The sensor was operated in IMUPLUS mode, providing an orientation estimate from on-chip fusion of gyroscope and accelerometer measurements. The Arduino transmitted yaw heading, encoder, and obstacle-distance measurements to the Jetson through a serial connection at approximately 20~Hz. Communication was bidirectional, with the Jetson also sending auxiliary commands for an onboard display and a buzzer. 

A 16$\times$2 LCD screen and an active buzzer provided immediate feedback during field experiments (Figure~\ref{fig:diagram_robot}). The LCD displayed the encoder signal and compass heading, together with the GNSS RTK status, number of satellites, and estimated horizontal positioning accuracy. This allowed the operation of the encoders, compass, and GNSS receiver, as well as the quality of the GNSS position estimate, to be checked directly on the robot. 
A 5V buzzer provided audible confirmation when a visual snapshot was successfully acquired. The buzzer also confirmed when CSV recording started at the beginning of an experimental trial.

Visual input was acquired using a Kodak PIXPRO SP360 4K panoramic RGB camera mounted at the top of the robot (Figure~\ref{fig:robot} and Figure~\ref{fig:diagram_robot}). The camera uses a fixed-focus \(235^\circ\) fisheye lens, producing a circular panoramic image that was subsequently unwrapped into a rectangular representation for visual processing.

\subsection{Visual processing and visual compass model} Camera frames were acquired continuously in a separate thread, while the 20~Hz control loop used the latest available frame to generate visual-compass predictions. The current frame was resized to $100\times100$ pixels before being passed through the visual pipeline. The fisheye image was unwrapped to a $50\times50$ panoramic representation, vertically cropped to $50\times28$ pixels, reduced to the green channel, and processed using a $3\times3$ centre-surround lateral-inhibition filter. The resulting image was finally resized to $10\times10$ pixels and flattened to form the 100-element input to the visual-compass model. The visual compass consisted of 20,000 KCs with five visual inputs per KC, 1\% sparse activity, and eight directionally tuned MBONs. The implementation followed the model described in \citet{Misiek2026}.

\subsection{Path integration}

Path integration used the same abstract, CX inspired model used in \citet{Misiek2026}, with inertial heading measurements provided by the BNO055 and wheel-encoder information. In contrast to the simulations, no artificial heading or stride noise was added. The PI model was already exposed to substantial real-world noise, such as the evident slippage of the wheel against the grassy and uneven ground, during the experiments.

\subsection{Behavioural modes}

During the experiment, the robot was controlled using a Logitech F710 wireless gamepad, connected to the Jetson through its USB receiver. The buttons of the joystick triggered predefined commands, including activating manual control, PI homing, visual homing, starting a LW (autonomous sequence including spiralling behaviour intersped with pirouettes), start CSV recording, reset the PI vector, consolidate visual memories and create a new agent. (Figure~\ref{fig:diagram_robot})

\paragraph{Path integration homing}
PI homing mode used only the PI vector to perform homing.
During path-integration homing, the egocentric PI direction $\theta_{\mathrm{PI}}$ was used directly as a steering error. When
\begin{equation}
|\theta_{\mathrm{PI}}| > 20^\circ,
\end{equation}
the robot rotated in place with
\begin{equation}
\omega =
\left(
0.4
+
0.4\frac{|\theta_{\mathrm{PI}}|}{180^\circ}
\right)
\mathrm{sign}(\theta_{\mathrm{PI}}),
\qquad
v=0.
\end{equation}
Once the angular error was within $20^\circ$, the robot moved forward with two third of its maximal power.

\paragraph{Visual homing}
A separate visual-homing mode used visual-compass predictions only to return to the nest. Because the visual compass produced an egocentric nest direction, this prediction was first transformed into an allocentric direction using the current BNO055 heading. This direction was then provided to a separate central-complex-inspired ring representation, which acted as a low-pass filter. 

The CX ring consisted of eight neurons with preferred directions spaced by $45^\circ$. For each allocentric visual-compass prediction $\phi_t$, the activity of neuron $i$ was updated according to
\[
a_i(t)=a_i(t-1)+g\cos(\theta_i-\phi_t),
\]
where $\theta_i$ is the neuron's preferred direction and $g$ is the update gain. The represented direction was then decoded from the population vector,
\[
\hat{\phi}_t=\operatorname{atan2}\left(\sum_i a_i\sin\theta_i,\sum_i a_i\cos\theta_i\right).
\]
Thus, successive visual-compass predictions were integrated into a single directional estimate, reducing the influence of transient prediction noise.

Without this integration step, visual-compass predictions were highly noisy, producing erratic steering movements that prevented stable navigation. The CX ring smoothed successive visual predictions into a more stable directional signal for steering.[]

Without this integration step, visual-compass predictions were highly noisy, producing erratic steering movements that prevented stable navigation. The CX ring smoothed successive visual predictions into a more stable directional signal for steering. This is conceptually similar to \citet{Wystrach2026}, where successive visual-memory signals are integrated in the CX to update a persistent goal-heading representation.

The direction represented by the CX ring (the visual-compass goal) was compared with the robot's current heading to compute an angular steering error. The robot then steered towards this goal, with rotational speed increasing and forward speed decreasing as the angular error increased.

During visual homing tests, the robot forward speed was of 0.495~m/s on average.

\paragraph{Spiraling behaviour}
During a LW, the visual compass prediction was compared with the current PI estimate. Their absolute angular disagreement was defined as
\begin{equation}
E_{\mathrm{PI/VC}}
=
\frac{
\left|
\mathrm{wrap}
\left(
\theta_{\mathrm{VC}}
-
\theta_{\mathrm{PI}}
\right)
\right|
}{180^\circ},
\end{equation}
such that $E_{\mathrm{PI/VC}}$ ranged from 0 for complete agreement to 1 for a $180^\circ$ disagreement.
When the PI and visual compass directions differed by less than $25^\circ$, the robot moved outward. Otherwise, the robot followed the spiral direction. The spiral direction was defined as the PI vector, offset by
\begin{equation}
\theta_{\mathrm{search}}
=
\theta_{\mathrm{PI}}
+
s \times 77^\circ,
\end{equation}
where $s\in\{-1,1\}$ determined the direction of the search. During this phase, the robot moved at a third of its maximum power, while continuously steering toward the desired direction. This mechanism allowed the robot to continuing spiralling around the nest until the visual-compass error decreased with accumulating experience. A 77$^\circ$ angle was chosen for spiralling instead of a 90$^\circ$ angle, to counteract for the residual outward motion observed in naive agent. 

\paragraph{Pirouettes}
The normalized PI--visual-compass disagreement was accumulated during spiralling behaviour. When the local accumulated error exceeded a value of 15 ($180^\circ$ error = 1), the robot initiated a pirouette. The robot stopped translating and started a pirouette. It rotated in place in a randomly selected direction (to the left or to the right), at half power. Visual snapshots were acquired during the rotation at a rate of approximately 1.3~Hz. Snapshots were captured while the robot rotated, without fixations. Captured images showed no visible motion blur (Fig.~\ref{fig:panoramic}), and the snapshot acquisition process was very fast, occurring on the order of tens of milliseconds, making it likely that the recorded view and heading closely matched those at the time the snapshot was triggered. Whether fixation periods would nevertheless improve subsequent homing performance remains to be tested.

Completion of a pirouette was detected when the robot moved returned to to within $25^\circ$ of the initial orientation before the pirouette. The error accumulator was then reset and spiralling behaviour resumed.

A second accumulator integrated PI--visual-compass error over the full LW. When this accumulated error exceeded a value of 200, the controller automatically switched to PI homing mode. 

PI continued to be integrated during pirouettes, and the opposite signal from the left and right front wheels encoders mostly cancelled out.

\subsection{Obstacle avoidance}

Obstacle detection was supported by two HC-SR04 ultrasonic range sensors mounted at the front of the robot, one on each side near the front wheels  (Figure~\ref{fig:robot} and Figure~\ref{fig:diagram_robot}). The two HC-SR04 were connected to the Arduino.
When a cylinder was detected within approximately 60~cm, the robot reversed to avoid a collision. 

\subsection{GNSS Tracking}

The robot spatial position was measured using an RTK-enabled u-blox ZED-F9P GNSS receiver mounted on the robot and connected to the Jetson  (Figure~\ref{fig:diagram_robot}). The GNSS antenna was mounted as close as possible to the center of the robot, 4~cm from the axis of rotation. Instead of using a local base station, RTK correction data were received over the internet from the public RTK2GO service using NTRIP. Corrections were provided by the \textit{Ashcombe Farm} base station, located approximately 4~km from the experimental site, and transmitted to the ZED-F9P receiver via an LTE modem.
The ZED-F9P applied these corrections to its GNSS measurements in real time, providing centimetre-scale positioning when an RTK signal was available, and approximately 0.5--2~m accuracy otherwise. Latitude, longitude, RTK status, satellite count, horizontal position accuracy, ground speed, and related GNSS measurements were recorded together with the robot's navigation state. GNSS positions were converted to local Cartesian easting and northing coordinates for subsequent trajectory analysis.

\subsection{Reliability of the GNSS signal during recordings}
\label{sec:GNSS_reliability}

\begin{figure}
\begin{adjustwidth}{-2.2in}{0in}
\centering    \includegraphics[width=\columnwidth,height=0.85\textheight,keepaspectratio]{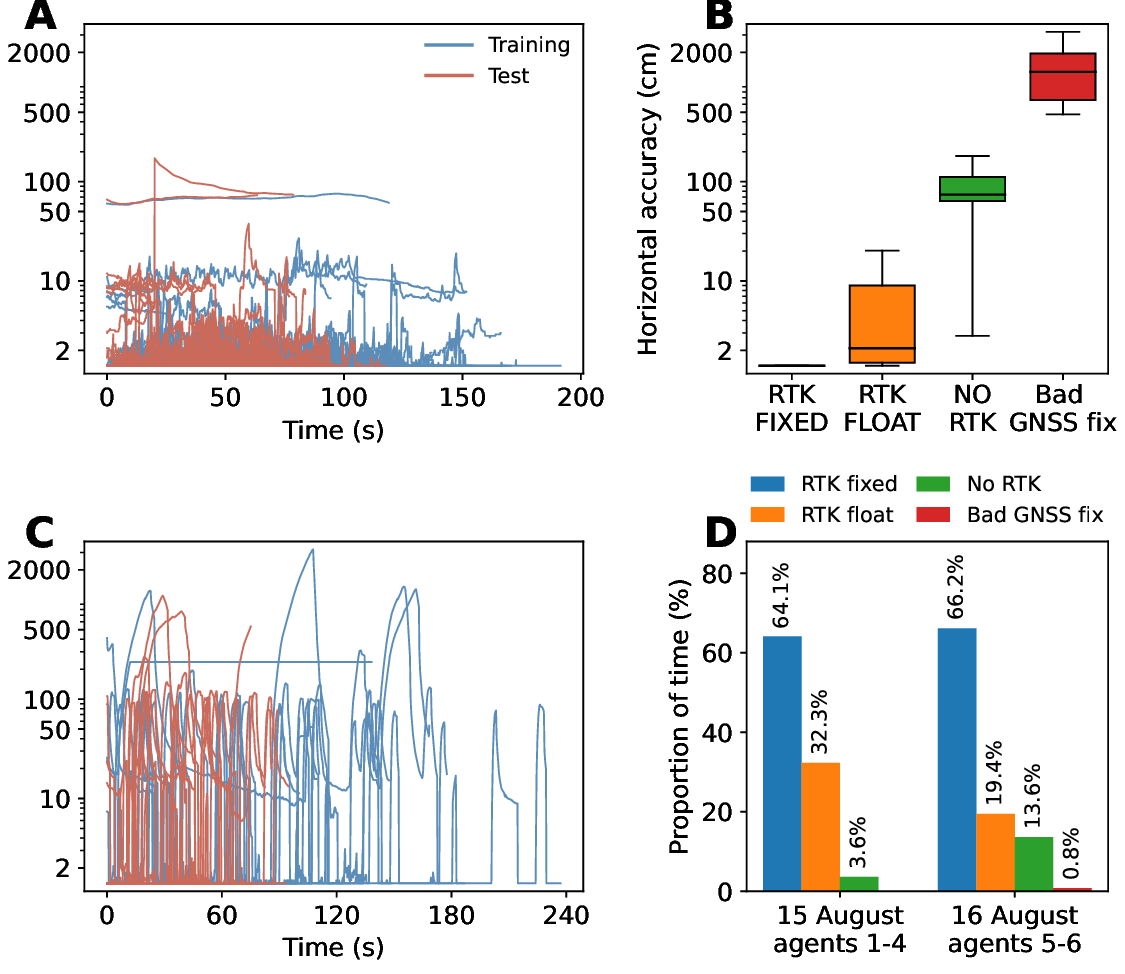}
\vspace{5mm}
\caption{
\textbf{GNSS horizontal accuracy.}
\textbf{(A)} Horizontal accuracy over time for agents 1--4, tested on 15 August 2026. Each line corresponds to a LW (blue) or visual homing test (red). Most measurements were below 5~cm and nearly all below 20~cm; the worst was approximately 2~m.
\textbf{(B)} Distribution of horizontal accuracy across the four GNSS statuses for agents 1--6. Blue (left flattened box): RTK fixed; orange: RTK float; green: no RTK; red: bad GNSS fix. Accuracy was high for RTK fixed and RTK float states, ranging from very precise in RTK fixed (~1.4~cm) to unusable in bad GNSS fix (>10m).
\textbf{(C)} Horizontal accuracy for agents 5 and 6, tested on 16 August 2026. Accuracy exceeded 2~m in four walks and even reached up to 33~m. One walk showed sustained poor accuracy above 2~m.
\textbf{(D)} Proportion of recording time spent in each GNSS status. Agents 1--4 (left), tested on 15 August, spent almost all recording time in RTK fixed or RTK float states. Agents 5--6 (right), tested on 16 August, spent more time without RTK and nearly 1\% of the data was completely unusable due to bad GNSS fix.
}
\label{fig:gps_precision}
\end{adjustwidth}
\end{figure}

Throughout our LWs and visual homing tests, we recorded the trajectories using GNSS. The GNSS module provides a measure of horizontal accuracy, so we could verify that the recorded data was accurate and remove low precision data when necessary. We found that the GNSS signal was more accurate on 15 August than on 16 August (Fig.~\ref{fig:gps_precision}A,C). 

For agents 1--4, tested on 15 August, mean horizontal accuracy was 4.89~cm and median accuracy was 1.40~cm. For agents 5 and 6, tested on 16 August, the median remained 1.40~cm, but short periods of substantially poorer GNSS accuracy increased the mean to 68.10~cm (Fig.~\ref{fig:gps_precision}C).

We analysed GNSS status across all trials (Fig.~\ref{fig:gps_precision}B). RTK fixed measurements had a mean of 26.15 satellites used and 1.43~cm horizontal accuracy, while RTK float measurements averaged 25.10 satellites and 8.99~cm. Without RTK correction, accuracy decreased to 152.64~cm, with 21.21 satellites on average. Bad GNSS fixes were substantially worse, with only 3.23 satellites used and 14.18~m horizontal accuracy (Fig.~\ref{fig:gps_precision}B).

Agents 1--4 spent only 3.6\% of recording time without RTK correction and never entered a bad GNSS fix. In contrast, agents 5 and 6 spent 13.6\% of recording time without RTK correction and 0.8\% in bad GNSS fix, corresponding to brief periods of very poor accuracy with errors reaching 33~m (Fig.~\ref{fig:gps_precision}D). Eventually, the cause of the degraded GNSS fixes on the second day remained unclear, but they coincided with a substantial loss of satellites and poorer satellite geometry. Datapoints with an estimated horizontal error greater than 1~m were removed from the analyses in the following result section, corresponding to 0.20\% of CSV rows for agents 1--4 and 7.86\% for agents 5--6.

\subsection{Analysis}
All analyses were performed from the CSV files recorded onboard during each LW or test, which contained the robot's navigation state together with measurements from the BNO055, wheel encoders and ZED-F9P GNSS receiver. Data were recorded at 10~Hz, corresponding to one CSV row every 100~ms. Position was analysed from the ZED-F9P coordinates converted to local easting and northing, while forward speed was taken directly from the GNSS ground-speed estimate. For analyses requiring the robot's allocentric heading, the fast BNO055 yaw signal was anchored to the GNSS direction of travel: the GNSS--BNO angular offset was estimated and slowly updated during reliable, approximately straight forward motion, and held fixed during rotations and pirouettes, when GNSS course is unreliable. Path length was obtained by summing successive GNSS displacements, and path straightness was defined as the maximum displacement from the starting position divided by the total distance travelled. Unfortunately, pitch and roll data from the BNO055 were not recorded during the experiments and could therefore not be used to assess camera stabilisation.

\bibliography{references}

\appendix
\label{app}

\section{Supplementary analyses of the robot experiments}
\label{app:robot}

\subsection{Correlation between prediction variance and visual-compass accuracy}
\label{app:robot_variance}

\begin{figure}
    \centering
    \includegraphics[width=\columnwidth,height=.75\textheight,keepaspectratio]{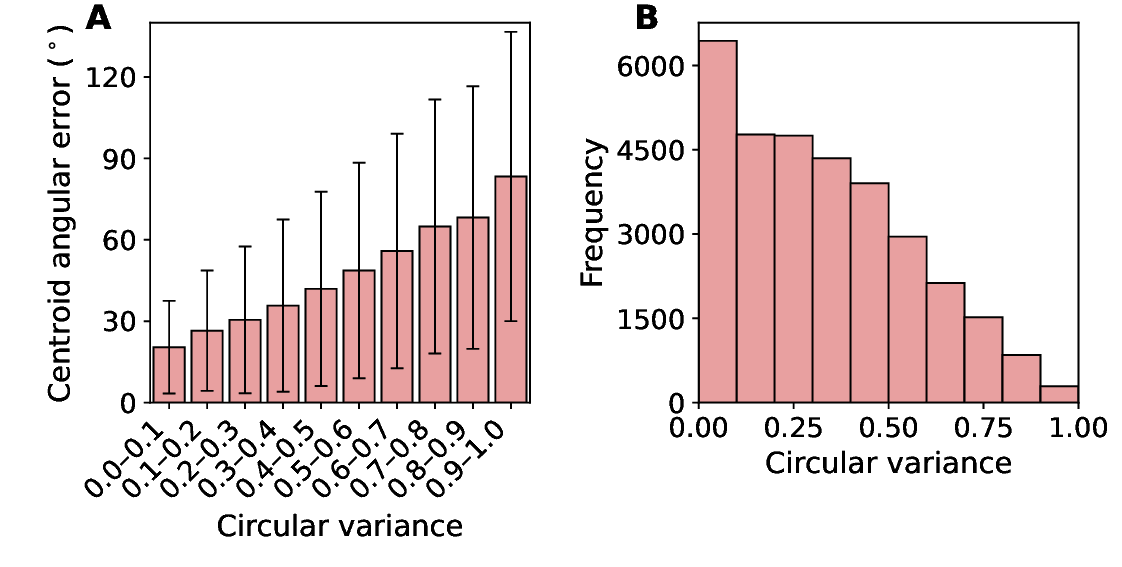}
    \caption{
    \textbf{Relationship between prediction variance and visual-compass accuracy on the robot.}
    \textbf{(A)} Absolute angular error of the centroid of 10 consecutive raw visual-compass predictions (1~s of sampling), relative to the ground-truth nest direction derived from GNSS position. During these 1-s windows, the maximum angular spacing between headings was $54.31^\circ \pm 20.63^\circ$ (mean $\pm$ SD), with a maximum of approximately $100^\circ$, while the maximum linear spacing between sampled positions was $0.287 \pm 0.212$~m, with a maximum of approximately $0.8$~m. Only LW3 and subsequent LWs were included, excluding LW1 and LW2. Mean centroid error increased monotonically with circular variance, from $20.42^\circ$ in the lowest-variance bin ($0$--$0.1$) to $83.32^\circ$ in the highest-variance bin ($0.9$--$1.0$). Prediction variance was positively correlated with centroid error (Spearman's $\rho = 0.358$, $p < 0.001$), with a linear $R^2 = 0.167$ and mutual information of $0.113$~nats. Error bars show standard deviations, which increased with variance. The variability in all variance bin was quite high, hence the low correlation and low predictability despite a clear trend on average.
    \textbf{(B)} Frequency distribution across all the one-second prediction samples across circular-variance bins. Low-variance samples were more frequent.
    The positive correlation between prediction variance and angular error suggests that the variance of successive visual-compass predictions could serve as a proxy for prediction accuracy, providing a confidence or reliability signal. This could be used to weight visual-compass predictions against PI, or to distinguish familiar from unfamiliar locations. Unlike the visual familiarity or novelty signals commonly derived from MBON activation in visual-compass models~\citep{Narendra2017,Sun2020,Amin2025-pm, Xiang2025}, this signal would reflect the consistency of successive directional predictions rather than the familiarity of the visual input itself.
    }
    \label{fig:robot_variance_corr}
\end{figure}

\end{document}